\documentclass[10pt,twocolumn,letterpaper]{article}

\usepackage[pagenumbers]{iccv} 

\definecolor{iccvblue}{rgb}{0.21,0.49,0.74}
\usepackage[pagebackref,breaklinks,colorlinks,allcolors=iccvblue]{hyperref}

\usepackage{multirow}
\usepackage{bm}
\usepackage{color, colortbl}
\definecolor{gray}{gray}{0.85}

\def\paperID{9282} 
\def\confName{ICCV}
\def\confYear{2025}

\title{SOAF: Scene Occlusion-aware Neural Acoustic Field}

\author{
  Huiyu Gao\textsuperscript{\textnormal{1}}, 
  Jiahao Ma\textsuperscript{\textnormal{1,2}}, 
  David Ahmedt-Aristizabal\textsuperscript{\textnormal{2}}, 
  Chuong Nguyen\textsuperscript{\textnormal{2}}, 
  Miaomiao Liu\textsuperscript{\textnormal{1}}\\
  \textsuperscript{\textnormal{1}}Australian National University, \textsuperscript{\textnormal{2}}CSIRO Data61 \\
   {\tt\small \{huiyu.gao, jiahao.ma, miaomiao.liu\}@anu.edu.au}\\ {\tt\small\{jiahao.ma, david.ahmedtaristizabal, chuong.nguyen\}@data61.csiro.au}
}

\begin{document}

\maketitle
\begin{abstract}
This paper tackles the problem of novel view audio-visual synthesis along an arbitrary trajectory in an indoor scene, given the audio-video recordings from other known trajectories of the scene.~Existing methods often overlook the effect of room geometry, particularly wall occlusions on sound propagation, making them less accurate in multi-room environments.~In this work, we propose a new approach called Scene Occlusion-aware Acoustic Field (SOAF) for accurate sound generation. Our approach derives a global prior for the sound field using distance-aware parametric sound-propagation modeling and then transforms it based on the scene structure learned from the input video. We extract features from the local acoustic field centered at the receiver using a Fibonacci Sphere to generate binaural audio for novel views with a direction-aware attention mechanism.
Extensive experiments on the real dataset~\emph{RWAVS} and the synthetic dataset~\emph{SoundSpaces} demonstrate that our method outperforms previous state-of-the-art techniques in audio generation.
\end{abstract}
\section{Introduction}
\label{sec:intro}

We live in a world with rich audio-visual multi-modal information. Audio-visual scene synthesis enables the generation of videos and corresponding spatial audio along arbitrary novel camera trajectories based on a source video with its associated binaural audio. This task involves reconstructing the audio-visual scene both visually and acoustically from audio-visual recordings of known trajectories. Specifically, it enables synthesizing the images a person would see and the sounds that a person would hear from any novel positions and directions while navigating within the scene. 

Neural Radiance Fields (NeRF)~\cite{nerf} has made significant progress in the field of computer vision over the past few years. NeRF uses Multi-Layer Perceptrons (MLP) to learn an implicit and continuous representation of the visual scene and synthesize novel view images through volume rendering. While NeRF has been extensively explored in the field~\cite{fridovich2022plenoxels, chen2022tensorf, muller2022instant, reiser2021kilonerf, yu2021plenoctrees, martin2021nerf}, these methods focus solely on the visual aspect of the input video, ignoring the accompanying audio track. However, the world we live in contains multi-modal information. Most videos we capture include not only visual images but also sound signals. Therefore, investigating novel view acoustic synthesis is crucial to providing more immersive experiences for users in various AV/VR applications. 
\begin{figure}[t]
\centering
\vspace{0.1cm}
\includegraphics[width=0.96\linewidth]{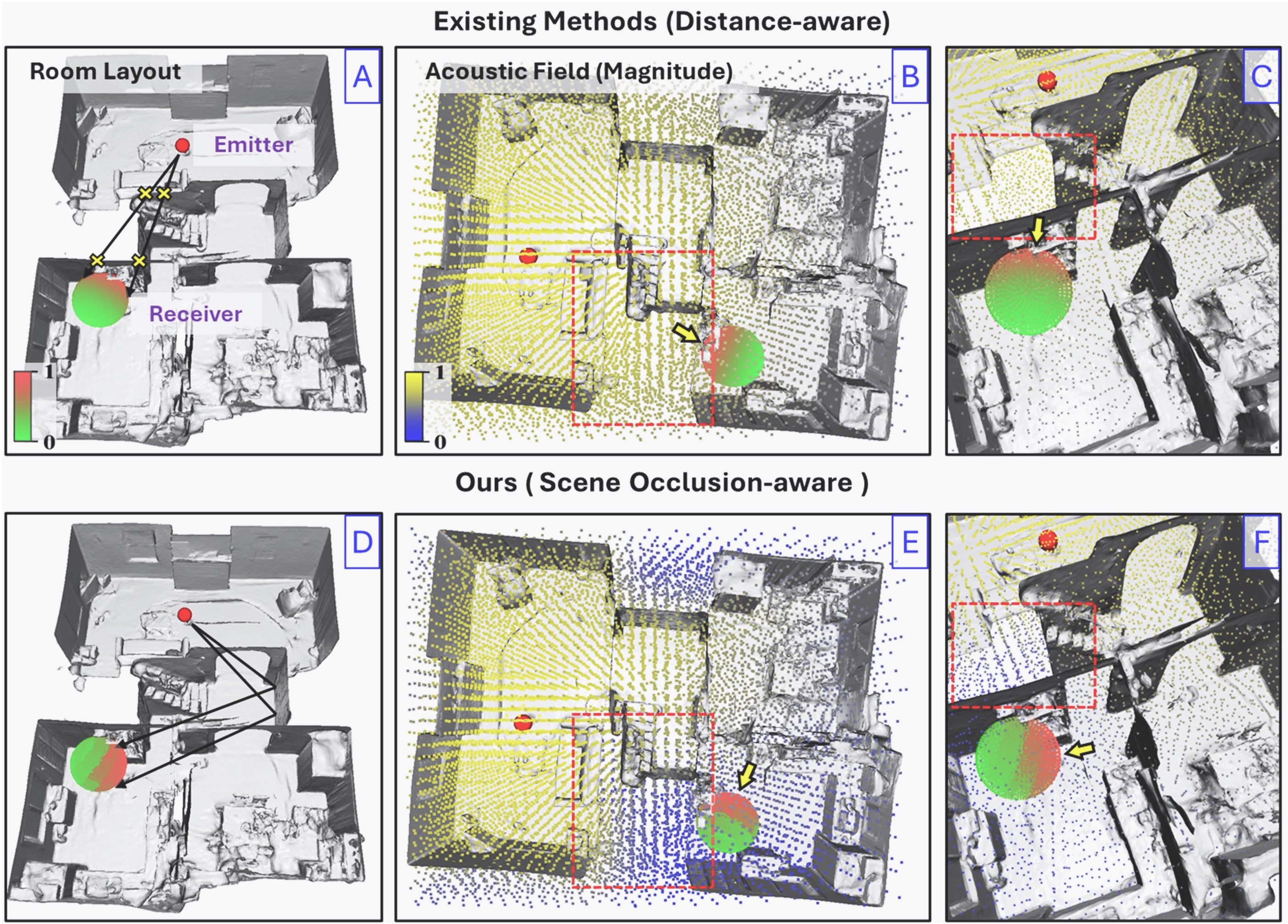}
\vspace{-0.1cm}
\caption{Pure distance-aware acoustic field~\cite{NAF,AVNeRF} \textit{vs.} our proposed Scene Occlusion-aware Acoustic Field (SOAF). \textbf{Left column} (A \& D) shows sound propagation in a room: the small ball represents the emitter and the large red-green ball represents the receiver. Coded colors indicate sound intensity, with red to green denoting high to low.
\textbf{Middle column} (B \& E) visualizes the magnitude distribution of the acoustic field, with yellow to blue indicating high to low. The comparison of sound attenuation through walls, highlighted by red dashed bounding boxes in sub-figures B, C, E and F emphasizes our consideration of wall obstruction. 
\textbf{Right column} (C \& F) highlights the existing methods' neglect of obstruction in sound propagation in C while the proposed method gives higher sound intensity near the door than near the wall in F. }
\label{fig:intro_fig}
\end{figure}

Recently, Neural Acoustic Field (NAF)~\cite{NAF} became the first work to explore the application of implicit representation in sound field encoding. Similar to NeRF, NAF uses an MLP to learn a continuous function of the neural acoustic field, optimizing it by supervising the generated Room Impulse Response (RIR) in the time-frequency domain at different emitter-listener location pairs and view directions. For the audio-visual scene synthesis task, AV-NeRF~\cite{AVNeRF} is the first multi-modal approach to address this problem. They utilize NeRF~\cite{nerf} for novel view synthesis and integrate the rendered novel view image and depth map as visual and geometric cues into audio generation. While AV-NeRF~\cite{AVNeRF} has demonstrated promising results using multi-modal data, it only renders a single image and depth map for the novel view, providing limited visual and geometric information about the scene. Moreover, as illustrated in Figure~\ref{fig:intro_fig}, previous methods~\cite{NAF, AVNeRF} do not fully explore the influence of scene geometry, particularly wall occlusions, on sound propagation. 
In this work, we further leverage visual inputs to infer complete room geometry, thereby introducing effective distance-aware and occlusion-aware priors for acoustic field learning, leading to improved audio synthesis.

More specifically, as the sound wave propagates through space, the sound intensity attenuates over distance and is reflected off, absorbed, or transmitted by surfaces~\cite{kuttruff2016room, garrett2020attenuation}. The intensity of the sound wave received at a 3D position is determined by the full 3D scene geometry and materials~\cite{chen2020soundspaces, li2018scene, tang2022gwa, ratnarajah2022mesh2ir}. 
Therefore, given the video sequences, we first learn a NeRF~\cite{nerf, wang2021neus, yu2022monosdf} from the visual images to obtain relevant information about the environment. 
To enhance the sound field modeling, we derive a scene occlusion-aware prior, termed the~\emph{global acoustic field}, based on simplified distance-aware parametric sound propagation modeling centered at the sound source and transformed by the learned scene structure from NeRF. We then extract the acoustic feature from the~\emph{local acoustic field} around the receiver using a Fibonacci Sphere, followed by a direction-aware attention mechanism to obtain binaural features. These features are used to generate binaural audio at novel views, demonstrating superior performance.

In summary, our contributions are as follows: 
(i) We introduce a scene occlusion-aware global prior for the sound field, enabling us to explicitly model scene geometry and occlusions for accurate audio generation. 
(ii) Our direction-aware attention mechanism effectively captures useful local features to enhance binaural audio generation.
Extensive experiments on real and synthetic datasets, including RWAVS and SoundSpace, demonstrate the superior performance of our approach compared to existing methods.

\begin{figure*}[t]
\centering
\includegraphics[width=1\linewidth]{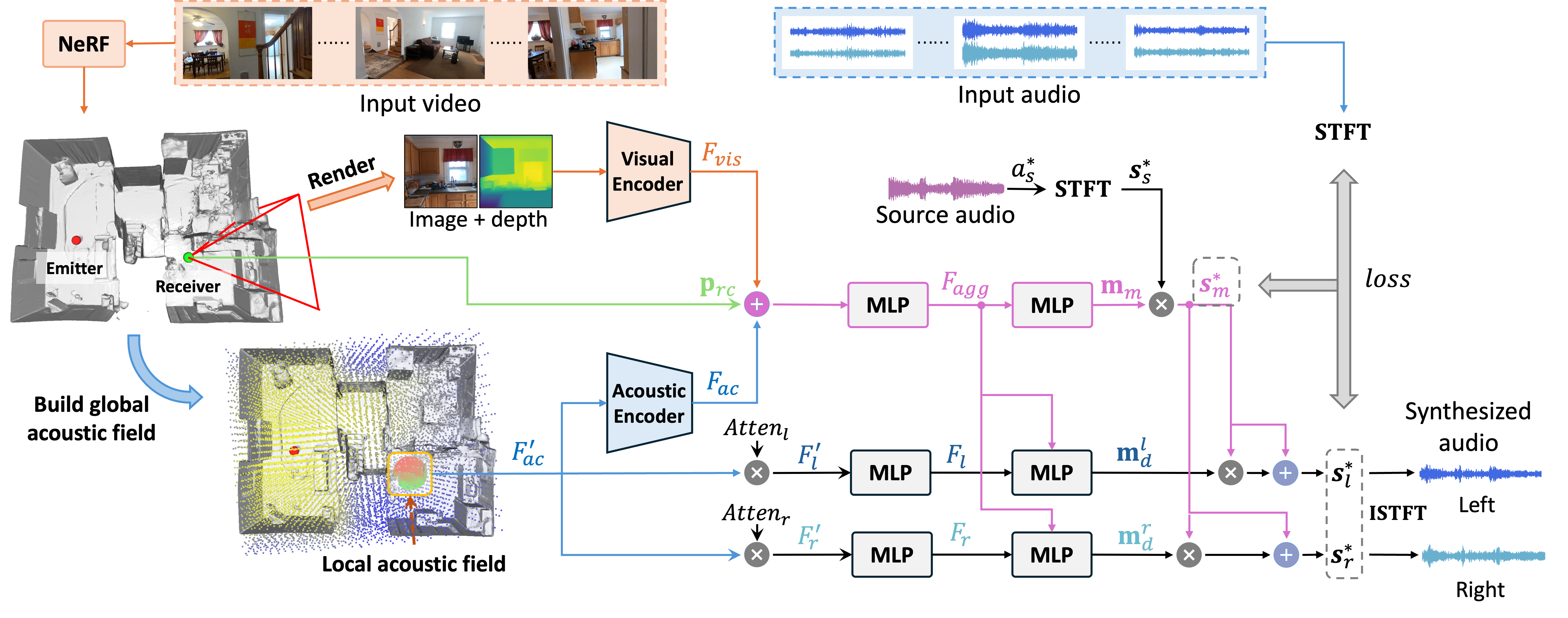}
\vspace{-0.66cm}
\caption{The pipeline of our proposed SOAF method. We first reconstruct the scene using NeRF from a calibrated video and build the global acoustic field. For audio synthesis at a new receiver pose $\bf p_{rc}$, we extract the acoustic feature $F_{ac}$ from the local acoustic field around the receiver, and combine it with $F_{vis}$ obtained from synthesized novel view images and $\bf p_{rc}$ to predict $F_{agg}$ and mixture acoustic mask ${\bf m}_m$. To distinguish left and right channels, we propose a direction-aware attention mechanism to generate channel-specific features $F'_l, F'_r$ based on distinct attention $Atten_l, Atten_r$ to the local acoustic field. Then the difference masks ${\bf m}_d^l$ and ${\bf m}_d^r$ are estimated with $F_{agg}$ combined with the refined channel feature $F_l$ or $F_r$, separately. Finally, we synthesize the binaural audio by combining the source audio magnitude ${\bf s}^*_s$ and the predicted masks ${\bf m}_m$, ${\bf m}_d^l$ and ${\bf m}_d^r$.}
\label{fig:pipeline}
\end{figure*}

\section{Related Work}

\textbf{Neural Radiance Fields and Implicit Surface.} 
NeRF~\cite{nerf} has emerged as a promising representation of scene appearance and has been widely used in novel view synthesis. Subsequent works~\cite{fridovich2022plenoxels, chen2022tensorf, muller2022instant, reiser2021kilonerf, yu2021plenoctrees, martin2021nerf} extend NeRF in various aspects, including faster training~\cite{fridovich2022plenoxels, chen2022tensorf, muller2022instant}, faster inference~\cite{reiser2021kilonerf, yu2021plenoctrees}, and handling in-the-wild images~\cite{martin2021nerf}. However, these methods struggle to extract high-quality surfaces due to insufficient surface constraints during optimization. To solve this issue, NeuS~\cite{wang2021neus} and VolSDF~\cite{yariv2021volume} propose utilizing the signed distance function (SDF) as an implicit surface representation and develop new volume rendering methods to train neural SDF fields. Some following works like MonoSDF~\cite{yu2022monosdf} and NeuRIS~\cite{wang2022neuris}, demonstrate the effectiveness of incorporating monocular depth priors~\cite{yu2022monosdf} and normal priors~\cite{yu2022monosdf, wang2022neuris} as additional geometric cues for learning implicit surface representation of indoor scenes from sequences of scene images. 
In our work, we follow~\cite{wang2021neus, yu2022monosdf} to achieve surface reconstruction from input images for occlusion-aware audio generation.

\noindent\textbf{Acoustic Fields.} 
The representation of spatial sound fields has been studied extensively. Previous methods have either directly approximated acoustic fields with handcrafted priors~\cite{mignot2013low, antonello2017room, ueno2018kernel} or focused solely on modeling perceptual cues with a parametric representations~\cite{raghuvanshi2014parametric, raghuvanshi2018parametric, chaitanya2020directional}. However, these methods often rely on strong assumptions.
Recently, researchers shifted towards data-driven sound fields learning. For instance, NAF~\cite{NAF} is the first method to leverage implicit representation to learn a neural acoustic field of RIR via an MLP. Unlike previous methods~\cite{mignot2013low, antonello2017room, ueno2018kernel, raghuvanshi2014parametric, raghuvanshi2018parametric, chaitanya2020directional} that capture scene acoustics with handcrafted parameterizations, implicit representation encodes the scene acoustics in a more generic manner, enabling application to arbitrary scenes. 
INRAS~\cite{INRAS} extends NAF by learning disentangled features for the emitter, scene geometry, and listener with known room boundaries. It reuses scene-dependent features for arbitrary emitter-listener pairs to generate higher fidelity RIR. 
DiffRIR~\cite{wang2024hearing} proposed a differentiable RIR rendering framework to reconstruct the spatial acoustic characteristics given RIR recordings and planar reconstruction of the scene. AVR~\cite{NEURIPS2024_Lan} introduced acoustic volume rendering to capture sound wave propagation principle and enforce acoustic multi-view consistency.
In contrast, we propose to explore general room geometry and introduce a scene occlusion-aware sound intensity prior obtained from the video frames for audio generation.

\noindent\textbf{Audio-visual Learning.} 
Several recent works~\cite{zhao2018sound, tian2018audio, chen2020soundspaces, chen2021semantic, mo2022localizing, gao20192, chen2022visual, somayazulu2024self, chen2023learning, chowdhury2023adverb, guo2021ad, zhou2022audio, zhou2020sep, ye2024lavss} have explored learning acoustic information from multimodal data for different tasks, including sound localization~\cite{tian2018audio, mo2022localizing}, audio-visual navigation~\cite{chen2020soundspaces, chen2021semantic}, visual-acoustic matching~\cite{chen2022visual, somayazulu2024self}, dereverberation~\cite{chen2023learning, chowdhury2023adverb}, and audio separation~\cite{zhou2020sep, ye2024lavss}. For novel view acoustic synthesis, ViGAS~\cite{ViGAS} combines auditory and visual observation from one viewpoint to render the sound received at the target viewpoint, assuming the sound source in the environment is visible in the input image and limited to a few views for audio generation. NACF~\cite{liang2023neural} integrates multiple acoustic contexts into audio scene representation and proposes a multi-scale energy decay criterion for supervising generated RIR. Few-shotRIR~\cite{majumder2022few} introduces a transformer-based model to extract multimodal features from few-shot audio-visual observations and predicts RIR for the queried source-receiver pair with a decoder module.~BEE~\cite{chen2023everywhere} reconstructs audio from sparse audio-visual samples by integrating obtained visual feature volumes with audio clips through cross-attention and rendering sound with learned time-frequency transformations. AV-NeRF~\cite{AVNeRF} synthesizes novel view audio by leveraging visual features extracted from images rendered from the novel view. In contrast, our approach 
explicitly models the effect of scene structure on sound propagation approximately for more realistic spatial audio generation.
\section{Task Definition}
The task of audio-visual scene synthesis, originally proposed in~\cite{AVNeRF}, aims to synthesize visual images and binaural audios for arbitrary receiver, such as camera and binaural microphone, within a static environment $E$. Given scene observations $O = \{O_1, O_2, \ldots, O_N\}$, where $O_i$ consists of an image $I$ with binaural audio clip $a_t$ recorded from pose $\hat{\mathbf{p}}_{rc}= ({\bf p}_{rc}, {\bf d}_{rc})$, defined as the receiver position ${\bf p}_{rc} \in \mathbb{R}^3$ and direction ${\bf d}_{rc}\in \mathbb{R}^3$, mono source audio clip $a_s$ with sound source position ${\bf p}_{sr} \in \mathbb{R}^3$, our goal is to generate new binaural audio $a_t^*$ and novel view image $I^*$ from a new receiver pose $\hat{\mathbf{p}}_{rc}^*$ and source audio $a_s^*$. 
It can be formulated as
\begin{equation}
    (a_t^*, I^*) = f(\hat{\mathbf{p}}_{rc}^*, a_s^*| O, E),
\end{equation}
where $f$ denotes the synthesis function. Commonly, it could be presented with the acoustic mask~\cite{AVNeRF} or Room Impulse Response (RIR)~\cite{NAF, INRAS}. 
In this paper, we aim to introduce the effective \textit{geometric prior} into the input. 
Our design can be applied to model both synthesis functions. In Section~\ref{sec:method}, we present our approach to predicting the acoustic mask; an alternative version of predicting the room impulse response is provided in the supplementary material.
\section{Method}
\label{sec:method}
We first provide an overview of our pipeline in Section~\ref{method:pipeline}. Then, we include visual feature extraction in Section~\ref{method:feature_extract}, details of our main contribution: global-local acoustic field generation in Section~\ref{method:gl_acoustic_field}, and the direction-aware attention mechanism in Section~\ref{method:dir_atten}. At last, we present the learning objective in Section~\ref{method:objective}.

\subsection{Overview}
\label{method:pipeline}
We adopt the acoustic mask-based synthesis function introduced in AV-NeRF~\cite{AVNeRF} for binaural audio prediction. Specifically, the acoustic mask consists of ${\bf m}_m, {\bf m}^l_d, {\bf m}^r_d  \in \mathbb{R}^{F \times W}$, where $F$ represents the frequency bins and $W$ is the number of time frames. ${\bf m}_m$ captures changes in audio magnitude at the receiver position ${\bf p}_{rc}$ relative to the sound source position ${\bf p}_{sr}$, ${\bf m}^l_d$ and ${\bf m}^r_d$ characterize the changes for left and right channels of the binaural audio. Given the spectrogram magnitude of the input audio clip $a^*_s$ after applying the Short-Time Fourier Transform (STFT), defined as ${\bf s}^*_s = \text{STFT}(a^*_s) \in \mathbb{R}^{F \times W}$, and predicted acoustic masks ${\bf m}_m$, ${\bf m}^l_d$, ${\bf m}^r_d$, we can synthesize the changed magnitude of the binaural audio as
\begin{equation}
  \begin{aligned}
    {\bf s}_m^* &= {\bf s}^*_s \odot {\bf m}_m, \\
    {\bf s}_l^* &= {\bf s}_m^* \odot (1 + {\bf m}^l_d), \\
    {\bf s}_r^* &= {\bf s}_m^* \odot (1 + {\bf m}^r_d),
  \end{aligned}
\label{eq:acoustic_mask}
\end{equation}
where $\odot$ denotes element-wise multiplication operation, ${\bf s}_l^*$ and ${\bf s}_r^*$ represent the magnitude of the synthesized audio in the left and right channel, respectively. Finally, we can obtain the binaural audio as $ a_t^* = [\text{ISTFT}({\bf s}_l^*), \text{ISTFT}({\bf s}_r^*)]$ where \text{ISTFT} denotes the inverse STFT.

The overall pipeline of our work is shown in Figure~\ref{fig:pipeline}. Our framework synthesizes novel view images for visual features and extracts scene geometry information from the input video, then builds the global acoustic field and local acoustic field to obtain the auditory features for the mask predictions. We provide details below.

\subsection {Image Synthesis and Visual Feature Extraction}
\label{method:feature_extract}
Following AV-NeRF~\cite{AVNeRF}, we achieve novel view image synthesis and visual feature extraction by learning a NeRF from the input video. In particular, we render an RGB image and a depth map from the receiver's pose, then extract color and depth features from the rendered images with a pre-trained encoder of ResNet-18~\cite{he2016deep}. These extracted features are concatenated as the visual feature $F_{vis}$, which provides important visual cues about the environment.

\begin{figure}[!t]
\centering
\includegraphics[width=0.73
\linewidth]
{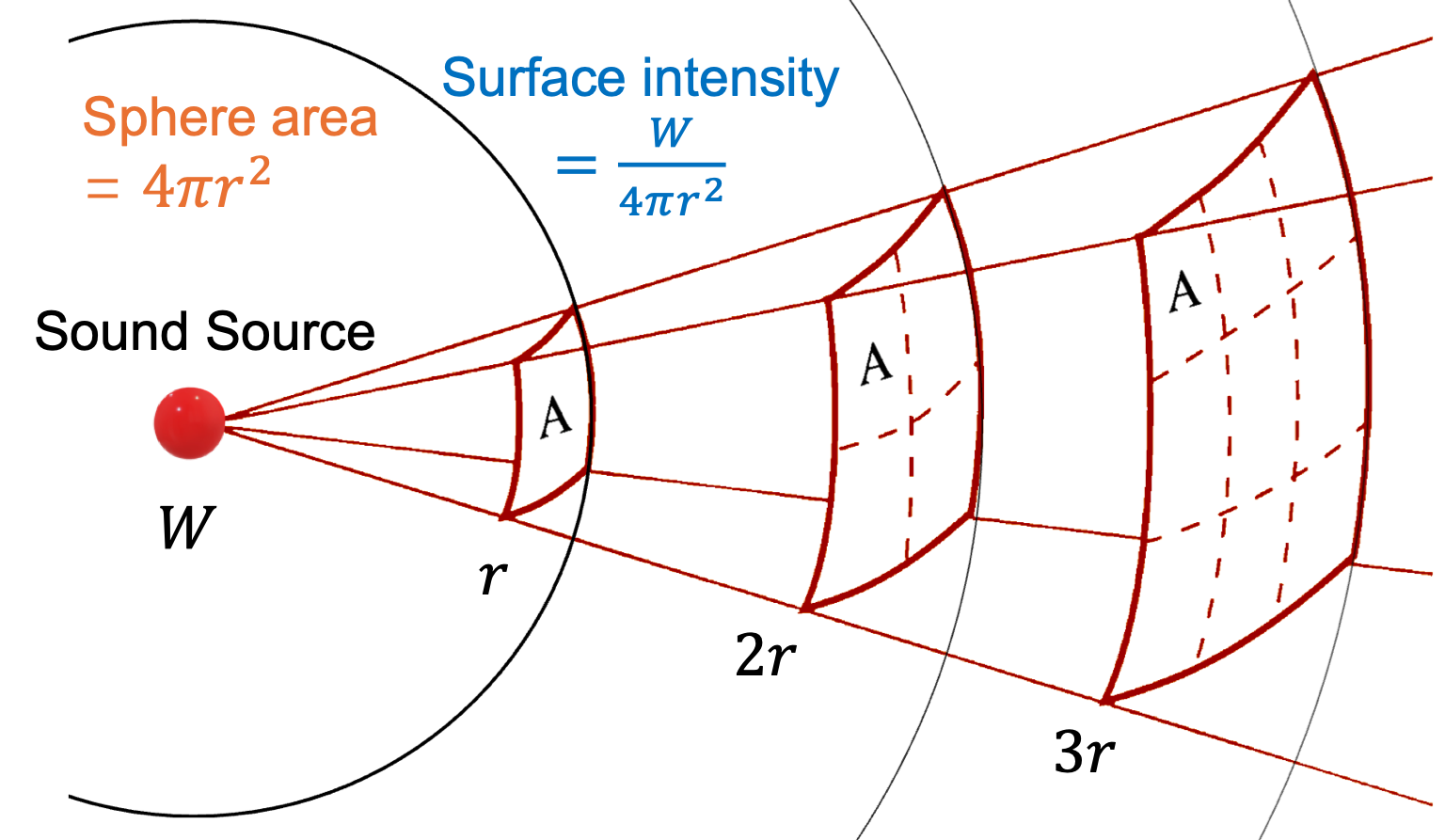}
\vspace{-0.1cm}
\caption{Illustration of the inverse square law~\cite{embleton1954mean, voudoukis2017inverse}. As the sound wave travels away from its source, the energy twice as far away from the source is distributed over four times the area, hence the intensity is one-quarter.}
\label{fig:inverse_square_law}
\end{figure}
\begin{figure}[!t]
\centering
\includegraphics[width=0.8
\linewidth]{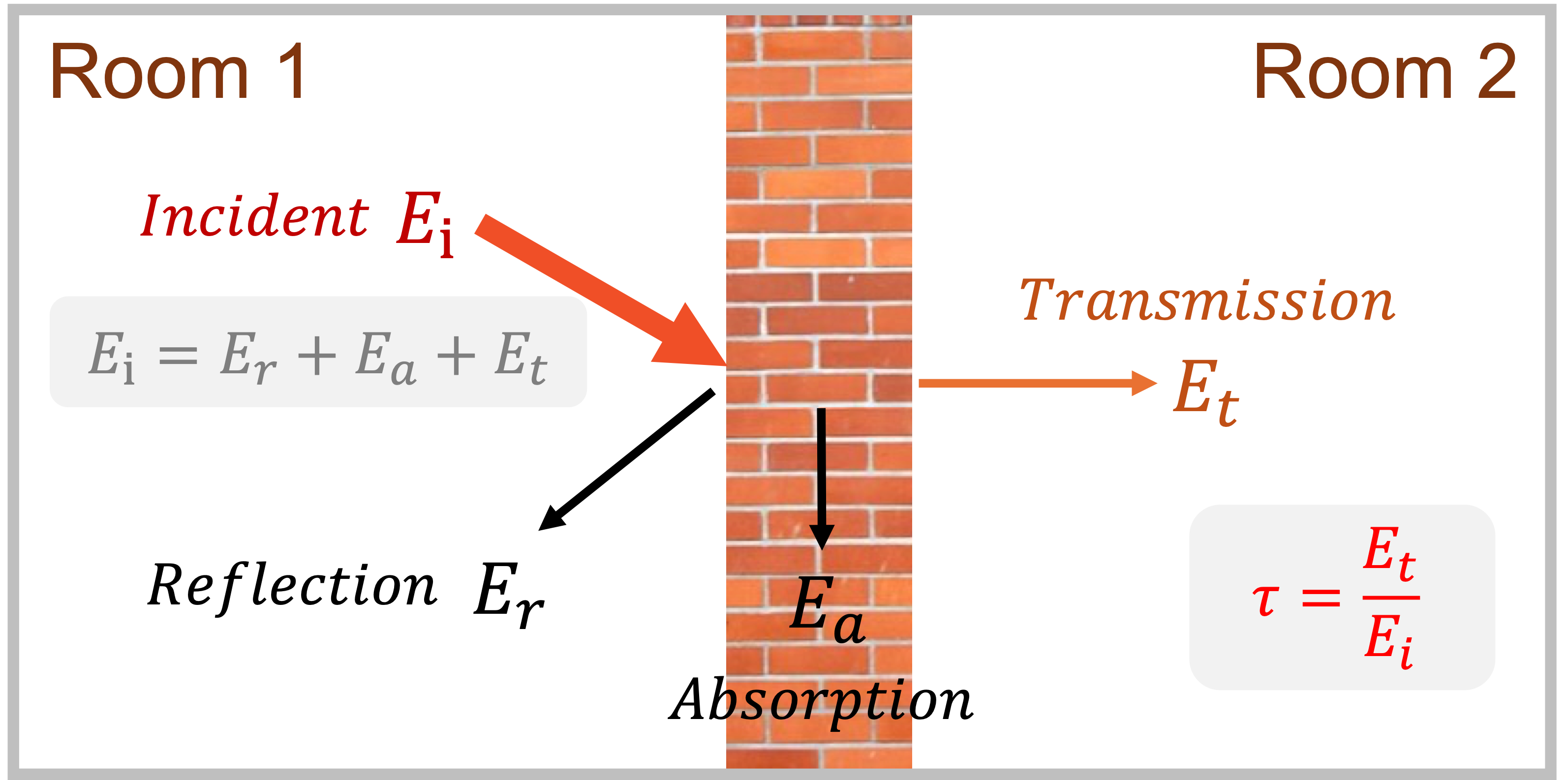}
\vspace{-0.1cm}
\caption{Illustration of the sound transmission coefficient $\tau$~\cite{bujoreanu2017experimental, tan2016sound}, which represents the ratio of transmitted sound energy when the sound wave travels through the barrier.}
\label{fig:transmit_wall}
\end{figure}

\subsection{Global and Local Acoustic Field}
\label{method:gl_acoustic_field}

\textbf{Global Acoustic Field.} The global acoustic field describes the distribution of sound intensity generated by the sound source throughout the scene. Since the 3D geometry and material properties determine sound propagation in an environment~\cite{chen2020soundspaces, li2018scene, tang2022gwa, ratnarajah2022mesh2ir, kuttruff2016room}, we derive a global prior of the acoustic field to enhance the audio synthesis quality based on the scene geometry extracted from the video frames. Given the sound source position ${\bf p}_{sr}$, we calculate the prior value at any point ${\bf p}_{i}$ in the scene by considering 
(i) the \textit{distance-aware} sound intensity attenuation; 
(ii) the \textit{occlusion-aware} sound intensity transmittance.

\textit{Distance-aware prior.} When sound propagates in the air, its intensity decreases as the distance from the sound source increases~\cite{garrett2020attenuation}. It obeys the \textit{inverse square law}~\cite{embleton1954mean, voudoukis2017inverse} in free space, which describes that the intensity of the sound is inversely proportional to the square of the distance from the sound source. As illustrated in Figure~\ref{fig:inverse_square_law}, the finite amount of energy created by the sound source is spread thinner and thinner along the expanding surface area of the sphere.~Inspired by this, we propose to explicitly quantify the sound attenuation based on the propagation distance, and calculate a prior value $E_0({\bf p}_i, {\bf p}_{sr})$ at the point ${\bf p}_i$ as
\begin{equation}
    E_0({\bf p}_{i}, {\bf p}_{sr}) = \frac{1}{4\pi d({\bf p}_{i}, {\bf p}_{sr})^2},
\end{equation}
where $d(\cdot, \cdot)$ denotes the Euclidean distance between points.

\textit{Occlusion-aware prior.} Unlike light waves, sound waves have much longer wavelengths, which allows them to diffract around small objects and propagate further~\cite{pasnau1999sound}. Large obstacles, particularly walls, significantly influence the acoustics in an indoor setting~\cite{long2005architectural}. Therefore, in this work, we check wall occlusions at the middle height of the scene to reduce the influence of small obstacles on the ground. To obtain the scene geometry, we learn a neural SDF field~\cite{Yu2022SDFStudio} from the visual images, and find the locations of scene walls with the zero-level set of SDF. As shown in Figure~\ref{fig:transmit_wall}, when sound waves travel through a wall from one side to another, the ratio of transmitted sound energy is termed as the \textit{sound transmission coefficient} $\tau$~\cite{bujoreanu2017experimental, tan2016sound}. It can be expressed as
\begin{equation}
    \tau = \frac{E_t}{E_i},
\end{equation} 
where $E_i$ and $E_t$ represent the sound energy before and after the sound waves traverse the barrier, respectively. 

Finally, for any point ${\bf p}_{i}$ in the scene, we generate a prior value $E({\bf p}_i, {\bf p}_{sr})$, which includes information about both propagation distance and wall occlusion effects, by
\begin{equation}
    E({\bf p}_i, {\bf p}_{sr}) = E_0({\bf p}_i, {\bf p}_{sr}) \times \tau^n, 
    \label{eq:prior}
\end{equation}
where $n$ is the number of walls found between the point ${\bf p}_i$ and the sound source ${\bf p}_{sr}$ in the SDF field. We convert $E({\bf p}_i, {\bf p}_{sr})$ to a logarithmic scale, following standard practice in acoustics, and normalize it to $\hat{E}({\bf p}_i, {\bf p}_{sr}) \in [0, 1]$ using the prior computed from maximum and minimum propagation distances across the scene.
Commonly, $\tau$ varies for different walls according to their material and thickness, here we simplify the case by choosing $\tau = 0.25$ for all walls~\cite{kuttruff2016room}. Note that we are not aim to perform precise sound propagation modeling, but rather to introduce informative priors that enhance the learning of the acoustic field.

\noindent\textbf{Local Acoustic Field.} The local acoustic field depicts the distribution of sound intensity around the receiver. Inspired by the design of spherical microphone arrays~\cite{spherical_microphone_array}, we utilize a Fibonacci Sphere~\cite{fibonacci_sphere} around the receiver to collect the sound intensity in the global acoustic field. 
Specifically, we first generate a unit Fibonacci Sphere with $G$ points on the sphere's surface centered at the receiver, then emit rays from the sphere center toward each surface point, in the directions ${\bf d}_{Fib}\in \mathbb{R}^{3 \times G}$. To build the local acoustic field, we uniformly sample $H$ points along each ray within the radius range $[r_{min}, r_{max}]$, resulting in a total of $G \times H$ sampled points. For each sampled point ${\bf p}_i$, we apply Equation~\ref{eq:prior} and obtain its prior $\hat{E}({\bf p}_i, {\bf p}_{sr})$. By integrating the prior values of the $H$ points along each direction, we obtain the local acoustic feature $F'_{ac} \in \mathbb{R}^{G}$, representing the priors across $G$ directions. Rather than treating all $H$ points equally along the same direction, we assign greater importance to points closer to the receiver by applying weights $w_i = e^{-d({\bf p}_{i}, {\bf p}_{rc})}\in [0, 1]$ and computing a weighted sum.

As shown in Figure~\ref{fig:pipeline}, our local acoustic feature $F'_{ac}$ is fed into an acoustic encoder to reduce the feature dimension and get the refined feature $F_{ac}$. Combining it with the visual feature $F_{vis}$ and the positional encoding of the receiver position ${\bf p}_{rc}$, we aggregate these features as $F_{agg}$ and estimate the mixture acoustic mask ${\bf m}_m$ with MLPs.

\subsection{Direction-aware Attention Mechanism}\label{method:dir_atten}

Given the local acoustic field, we propose a direction-aware attention mechanism to distinguish the left and right channel features to generate binaural audio. Specifically, we calculate the similarity between the left or right channel directions ${\bf d}_{l}, {\bf d}_{r}\in \mathbb{R}^3$ with the sphere points directions ${\bf d}_{Fib}$ to obtain the attention $Atten_{l}, Atten_{r} \in \mathbb{R}^{G}$ for each channel, then combine these attentions with local acoustic feature $F'_{ac}$ to obtain binaural features. It can be expressed as: 
\begin{equation}
Atten_{l} = {\bf d}_{Fib}^T {\bf d}_{l}, \quad
Atten_{r} = {\bf d}_{Fib}^T {\bf d}_{r},
\end{equation}
\begin{equation}
F'_{l} = F'_{ac} \odot Atten_{l}, \quad
F'_{r} = F'_{ac} \odot Atten_{r},
\end{equation}
where $\odot$ denotes element-wise multiplication. After further transformation of $F'_l$ and $F'_r$ to $F_l$ and $F_r$, respectively, to align their dimension with $F_{agg}$, we combine $F_{agg}$ with $F_l$ or $F_r$ separately to estimate ${\bf m}^l_d$ or ${\bf m}^r_d$.

\begin{figure}[!t]
\centering
\includegraphics[width=1\linewidth]{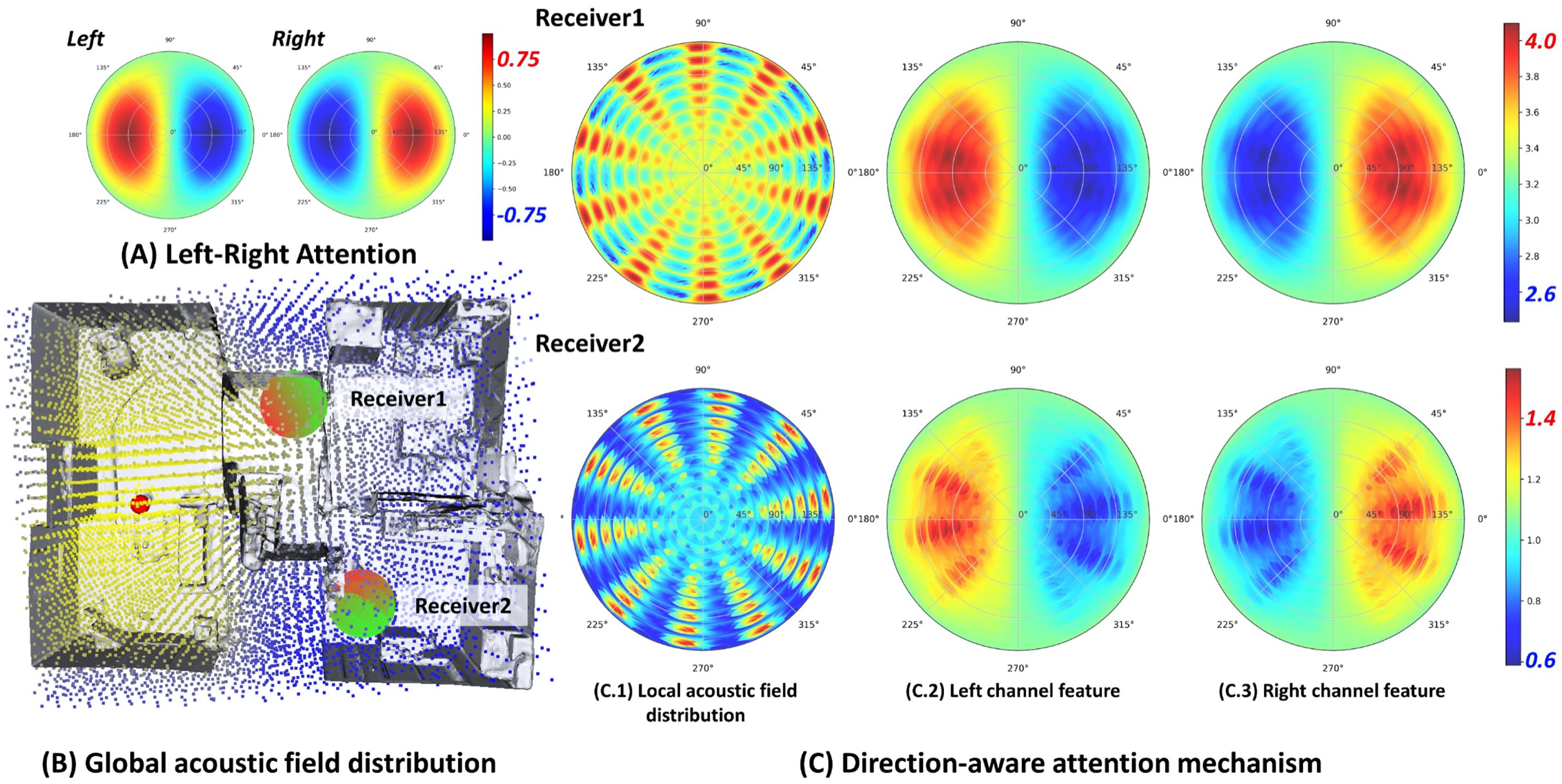}
\vspace{-0.6cm}
\caption{Direction-aware attention mechanism. (\textbf{A}) Predefined left-right attention. 
(\textbf{B}) Local distribution of two receivers: Receiver 1 in the hallway, close to the sound source with higher intensity; Receiver 2 in the kitchen, further away and obstructed with lower intensity. Intensity comparison is highlighted in the sub-figure C color bar. 
(\textbf{C}) The binaural features describe the spatial and directional sound characteristics generated by the combination of the left-right attention and the local acoustic field distribution.}
\vspace{-0.2cm}
\label{fig:direction_attention}
\end{figure}

Figure~\ref{fig:direction_attention} compares the local acoustic fields and binaural channel features for two receivers at different positions. The color bar shows the differing sound intensities of their left and right channels. Binaural features are generated by considering the channel directions and the sound intensity in the local sound field comprehensively.

\subsection{Learning Objective}
On the RWAVS dataset, we predict the acoustic masks ${\bf m}_m$, ${\bf m}^l_d$, ${\bf m}^r_d$ and obtain the estimated magnitudes ${\bf s}^*_m$, ${\bf s}^*_l$, ${\bf s}^*_r$ via Equation~\ref{eq:acoustic_mask}. Following the approach in~\cite{AVNeRF}, we optimize the network with the following loss function: 
\begin{equation}
    \mathcal{L}_A = \left\lVert {\bf s}_m -{\bf s}^*_m \right\rVert^2 + \left\lVert {\bf s}_l - {\bf s}^*_l \right\rVert^2 + \left\lVert {\bf s}_r - {\bf s}^*_r \right\rVert^2,
\end{equation}
where ${\bf s}_{m}$, ${\bf s}_{l}$ and ${\bf s}_{r}$ denote the ground-truth magnitudes, corresponding to the mixture, left, and right channels, respectively. The mixture ${\bf s}_m$ is defined as the average of ${\bf s}_{l}$ and ${\bf s}_{r}$. The first term of $\mathcal{L}_A$ encourages the network to predict the masks reflecting spatial effects caused by distance and geometry occlusion. The second and third terms encourage the network to generate masks that capture the differences between the left and right channels.
\label{method:objective}

\begin{table*}[t]
\centering
\resizebox{1\linewidth}{!}{
\footnotesize
\begin{tabular}{l|cc|cc|cc|cc|cc}
\toprule
\multirow{3}{*}{\quad Methods} & \multicolumn{2}{c|}{Office} & \multicolumn{2}{c|}{House} & \multicolumn{2}{c|}{Apartment} & \multicolumn{2}{c|}{Outdoors} & \multicolumn{2}{c}{\multirow{2}{*}{Mean}} \\ 
& \multicolumn{2}{c|}{w/o \textit{occ}} & \multicolumn{2}{c|}{w/ \textit{occ}} & \multicolumn{2}{c|}{w/ \textit{occ}} & \multicolumn{2}{c|}{w/o \textit{occ}} & \\
\cmidrule(lr){2-3} \cmidrule(lr){4-5} \cmidrule(lr){6-7} \cmidrule(lr){8-9} \cmidrule(lr){10-11}
& MAG$\downarrow$ & ENV$\downarrow$ & MAG$\downarrow$ & ENV$\downarrow$ & MAG$\downarrow$ & ENV$\downarrow$ & MAG$\downarrow$ & ENV$\downarrow$ & MAG$\downarrow$ & ENV$\downarrow$ \\
\midrule
Mono-Mono & 9.269 & 0.411 & 11.889 & 0.424 & 15.120 & 0.474 & 13.957 & 0.470 & 12.559 & 0.445\\
Mono-Energy & 1.536 & 0.142 & 4.307 & 0.180 & 3.911 & 0.192 & 1.634 & 0.127 & 2.847 & 0.160\\
Stereo-Energy & 1.511 & 0.139 & 4.301 & 0.180 & 3.895 & 0.191 & 1.612 & 0.124 & 2.830 & 0.159\\
\midrule
INRAS~\cite{INRAS} & 1.405 & 0.141 & 3.511 & 0.182 & 3.421 & 0.201 & 1.502 & 0.130 & 2.460 & 0.164\\
NAF~\cite{NAF} & 1.244 & 0.137 & 3.259 & 0.178 & 3.345 & 0.193 & 1.284 & 0.121 & 2.283 & 0.157 \\
ViGAS~\cite{ViGAS} & 1.049 & 0.132 & 2.502 & 0.161 & 2.600 & 0.187 & 1.169 & 0.121 & 1.830 & 0.150 \\ 
AV-NeRF~\cite{AVNeRF} & 0.930 & 0.129 & 2.009 & 0.155 & 2.230 & 0.184 & 0.845 & 0.111 & 1.504 & 0.145 \\
\rowcolor{gray} 
SOAF (Ours) & \textbf{0.795} & \textbf{0.125} & \textbf{1.952} & \textbf{0.153} & \textbf{2.098} & \textbf{0.182} & \textbf{0.737} & \textbf{0.108} & \textbf{1.396} & \textbf{0.142} \\
\bottomrule
\end{tabular}}
\vspace{-0.2cm}
\caption{Quantitative results on RWAVS dataset. Following the evaluation of AV-NeRF~\cite{AVNeRF}, our method consistently outperforms all baselines in MAG and ENV metrics across different environments. ``w/ \textit{occ}'' denotes a multi-room indoor scene with \textit{occ}lusion.}
\vspace{-0.2cm}
\label{tab:rwavs}
\end{table*}

\subsection{Implementation Details}
Our model is implemented using PyTorch \cite{paszke2019pytorch} and optimized by the Adam~\cite{adam} optimizer, with hyperparameters \(\beta_1 = 0.9\) and \(\beta_2 = 0.999\). The initial learning rate is set to \(5 \times 10^{-4}\) and is exponentially decreased to \(5 \times 10^{-6}\). The training process spans 200 epochs, with a batch size of 32. When building the Fibonacci Sphere, we set $G = 1024$ (uniformly generate 1024 rays around the sphere center) and $H = 10$ (uniformly sample 10 points along each ray), with the radius range $r_{min} = 0.01$  and $r_{max} = 1$. The acoustic transmission coefficient $\tau$ is set to 0.25. All our experiments are conducted on an RTX 4090 GPU.

\section{Experiments}

\begin{table}[t]
\centering
\resizebox{1\linewidth}{!}{
\footnotesize
\begin{tabular}{l|ccccc}
\toprule
Methods  & T60 (\%) $\downarrow$  & C50 (dB) $\downarrow$ & EDT (sec) $\downarrow$  \\
\hline
Opus-nearest  & 10.10 & 3.58 & 0.115 \\
Opus-linear & 8.64  & 3.13 & 0.097 \\
AAC-nearest & 9.35  & 1.67 & 0.059 \\
AAC-linear & 7.88  & 1.68 & 0.057 \\
\hline
NAF~\cite{NAF} & 3.18  & 1.06 & 0.031 \\
INRAS~\cite{INRAS} & 3.14  & 0.60 & 0.019 \\
AV-NeRF~\cite{AVNeRF} & 2.47 & 0.57 & 0.016\\
\rowcolor{gray} 
SOAF (Ours) & \textbf{2.29} & \textbf{0.54} & \textbf{0.014}\\
\bottomrule
\end{tabular}}
\vspace{-0.2cm}
\caption{Quantitative results on Soundsapces dataset. Following~\cite{AVNeRF, INRAS}, we evaluate the generated RIR using T60, C50, and EDT. Our method consistently performs the best in all metrics.}
\vspace{-0.2cm}
\label{tab:soundspaces}
\end{table}

\subsection{Datasets, Baselines \& Metrics}
\noindent\textbf{Datasets.} We evaluate our method on the real-world dataset RWAVS and the synthetic dataset SoundSpaces. 

\textit{RWAVS dataset.}
The Real-World Audio-Visual Scene (RWAVS) dataset is collected by the authors of AV-NeRF~\cite{AVNeRF} from diverse real-world scenarios, divided into four categories: \textit{office}, \textit{house}, \textit{apartment}, and \textit{outdoor} environments. In particular, the \textit{office} category includes indoor scenes with single-room layouts, while the \textit{house} and \textit{apartment} categories present scenes with multi-room layouts. To capture various acoustic and visual signals along different camera trajectories, the data collector moved randomly through the environment while holding the recording device. For each scene, RWAVS contains multimodal data including source audio, collected high-quality binaural audio, video, and camera poses, ranging from 10 to 25 minutes. Camera positions are densely distributed throughout the scene, and camera directions are sufficiently diverse. For a fair comparison, we maintain the same training/test split as~\cite{AVNeRF}, which contains 9,850 samples for training and 2,469 samples for testing, respectively.

\textit{SoundSpaces dataset.}
SoundSpaces~\cite{chen2020soundspaces, straub2019replica} is a synthetic dataset generated based on hybrid sound propagation methods~\cite{cao2016interactive, veach1995bidirectional, egan1989architectural} that simulates fine-grained acoustic properties by simultaneously considering the effects of room geometry and surface materials on sound propagation in a 3D environment. Following previous methods~\cite{NAF, INRAS, AVNeRF}, we validate our method on the same six representative indoor scenes, including two single rooms with rectangular walls, two single rooms with non-rectangular walls, and two multi-room layouts. For each scene, SoundSpaces provides binaural impulse responses for extensive emitter and receiver pairs sampled within the room at a fixed height from four different head orientations (0\textdegree, 90\textdegree, 180\textdegree, and 270\textdegree). To validate the effectiveness of our approach on this dataset, we modify our model to estimate binaural impulse responses instead of acoustic masks. We keep the same training/test split as previous works by using 90\% data for training and 10\% data for testing.

\begin{table*}[t]
\centering
\resizebox{0.98\linewidth}{!}{
\footnotesize
\begin{tabular}{l|cc|cc|cc|cc|cc}
\toprule
\multirow{3}{*}{\quad Methods} & \multicolumn{2}{c|}{Office} & \multicolumn{2}{c|}{House} & \multicolumn{2}{c|}{Apartment} & \multicolumn{2}{c|}
{Outdoors} & \multicolumn{2}{c}{Mean} \\
\cmidrule(lr){2-3} \cmidrule(lr){4-5} \cmidrule(lr){6-7} \cmidrule(lr){8-9} \cmidrule(lr){10-11}
& MAG$\downarrow$ & ENV$\downarrow$ & MAG$\downarrow$ & ENV$\downarrow$ & MAG$\downarrow$ & ENV$\downarrow$ & MAG$\downarrow$ & ENV$\downarrow$ & MAG$\downarrow$ & ENV$\downarrow$ \\ 
\midrule
Ours - w/o \textit{geo, dir} & 0.928 & 0.129 & 2.015 & 0.155 & 2.249 & 0.184 & 0.831 & 0.111 & 1.506 & 0.145 \\
Ours - w/o \textit{dir} & 0.868 & 0.127 & 1.977 & 0.154 & 2.168 & 0.183 & 0.785 & 0.109 & 1.450 & 0.143 \\
Ours - \textit{full} & \textbf{0.795} & \textbf{0.125} & \textbf{1.952} & \textbf{0.153} & \textbf{2.098} & \textbf{0.182} & \textbf{0.737} & \textbf{0.108} & \textbf{1.396} & \textbf{0.142} \\
\bottomrule
\end{tabular}
}
\vspace{-0.2cm}
\caption{Ablation study of proposed components on RWAVS dataset. ``\textit{geo}'' is global-local acoustic field, ``\textit{dir}'' is direction-aware attention mechanism and ``\textit{full}'' denotes all proposed modules.}
\vspace{-0.1cm}
\label{tab:rwavs_ablation}
\end{table*}
\begin{figure*}[!t]
\centering
\includegraphics[width=0.98\linewidth]{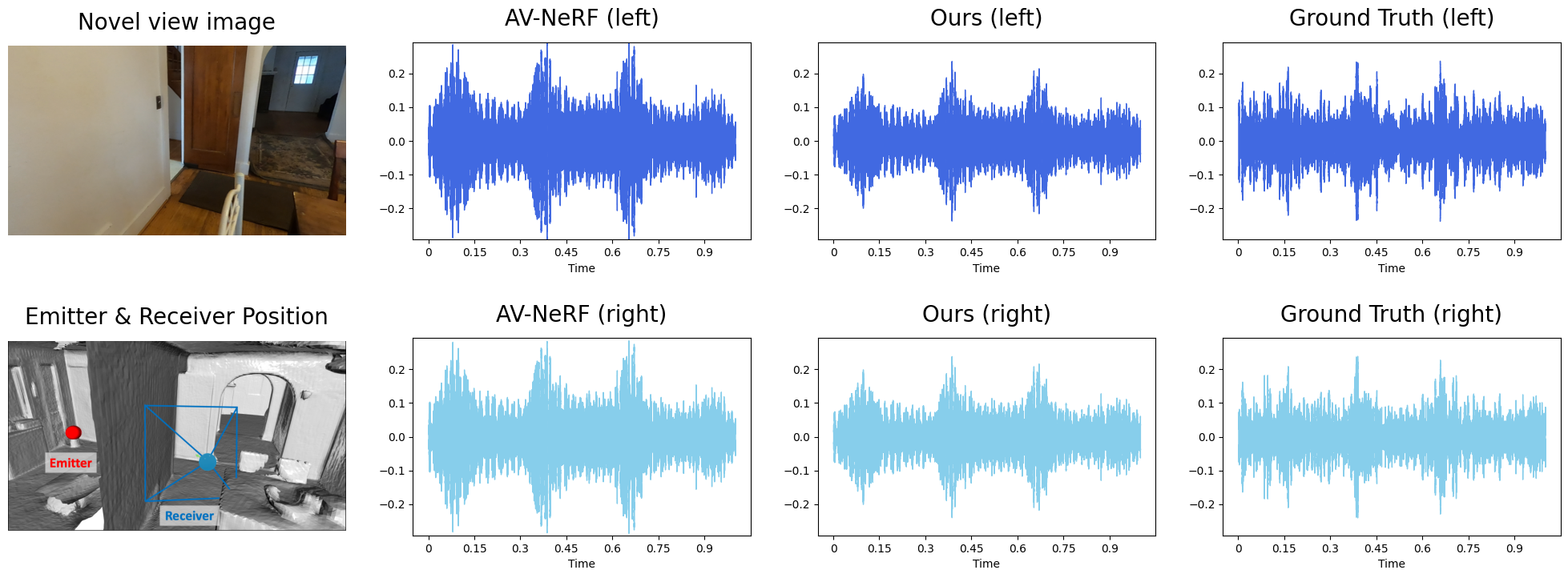}
\vspace{-0.3cm}
\caption{Example visual comparison of novel view audio synthesis. For the receiver, the emitter is blocked by a wall. Compared to AV-NeRF~\cite{AVNeRF} (MAG: 0.816; ENV: 0.124), our method (MAG: 0.471; ENV: 0.097) generates audio with more accurate energy attenuation.}
\vspace{-0.1cm}
\label{fig:vis}
\end{figure*}

\noindent\textbf{Baselines.} We compare our approach with state-of-the-art methods~\cite{NAF, INRAS, ViGAS, AVNeRF} that also learn a neural acoustic field with implicit representation. Among these methods, NAF~\cite{NAF} learns audio signals with a trainable local feature grid while INRAS~\cite{INRAS} disentangles scene-dependent features from audio signals and reuses them for all emitter-listener pairs. ViGAS~\cite{ViGAS} and AV-NeRF~\cite{AVNeRF} are multimodal approaches that leverage the visual feature of a single image for audio generation. 
For the RWAVS dataset, we include three additional baselines for reference: Mono-Mono, Mono-Energy, and Stereo-Energy. Mono-Mono simply repeats the source audio twice to achieve a binaural effect. Mono-Energy scales the energy of the source audio to match the average energy of the ground truth target audio then duplicates it to obtain a binaural audio. Stereo-Energy first duplicates the source audio and then scales the two channels separately to match the energy of each channel of the ground truth target audio. 
For SoundSpaces, we also compare our model with the linear and nearest neighbor interpolation results of two widely used audio coding methods: Advanced Audio Coding (AAC)~\cite{aac} and Xiph Opus~\cite{opus}. All these methods are evaluated with the same train/test split for each dataset.
\begin{figure*}[!t]
\centering
\vspace{-0.1cm}
\includegraphics[width=1\textwidth]{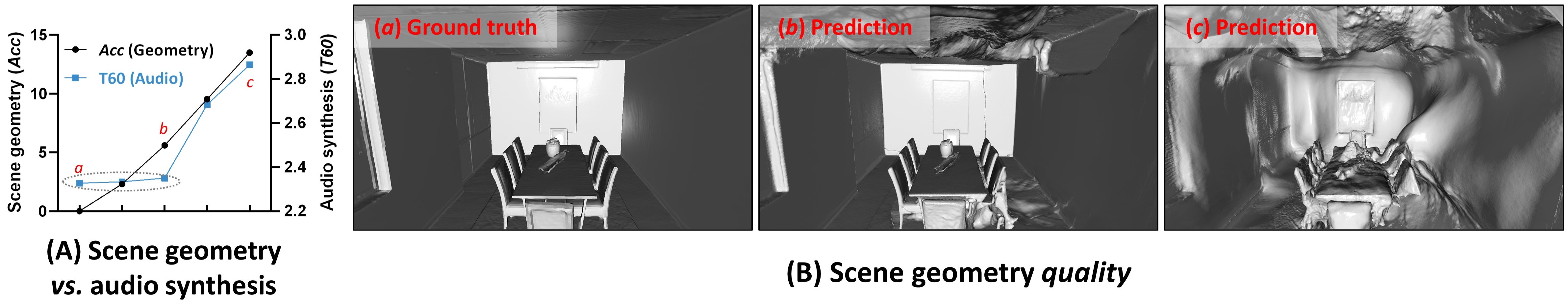}
\vspace{-0.66cm}
\caption{Robustness to scene reconstruction quality. \textbf{Left}: audio synthesis performance with various reconstructions. Small \textit{Acc} and \textit{T60} are better. \textbf{Right}: visual differences in reconstruction results. \textit{a}, \textit{b}, and \textit{c} correspond to the different geometry quality shown on the left.}
\vspace{-0.2cm}
\label{fig:ablation_geo}
\end{figure*}
\begin{figure*}[!t]
    \centering
    \vspace{-0.25cm}
    \small
    \renewcommand{\arraystretch}{0.4}
    \begin{tabular}{ccccc} \hspace{-0.2cm}
    \includegraphics[width=0.225\linewidth]{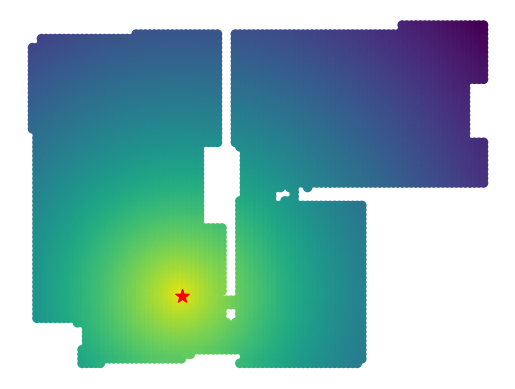} & \hspace{-0.2cm}
    \includegraphics[width=0.225\linewidth]{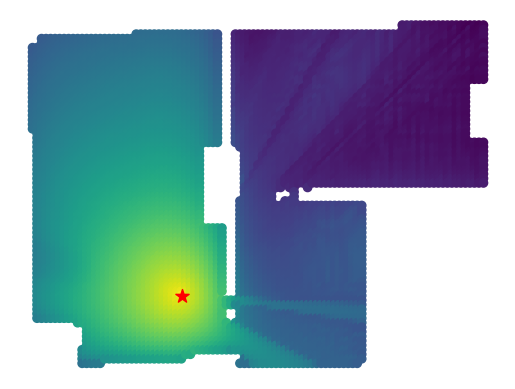} & \hspace{-0.2cm}
    \includegraphics[width=0.225\linewidth]{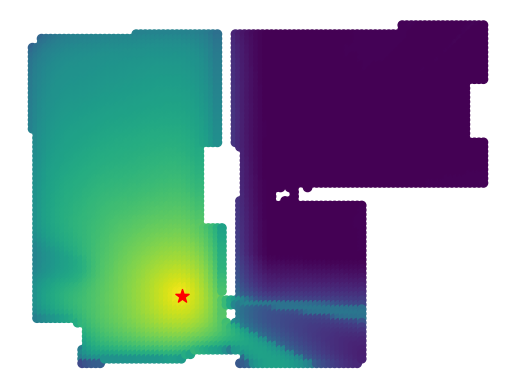} & \hspace{-0.2cm}
    \includegraphics[width=0.225\linewidth]{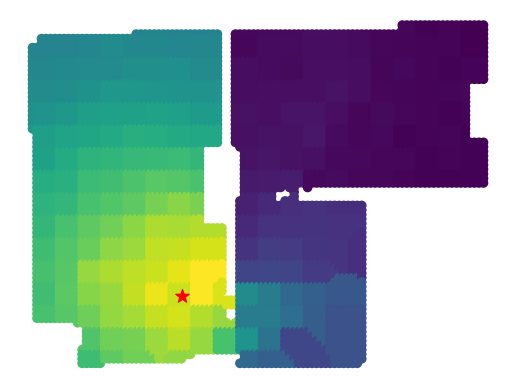} & \hspace{-0.3cm}
    \includegraphics[width=0.029\linewidth]{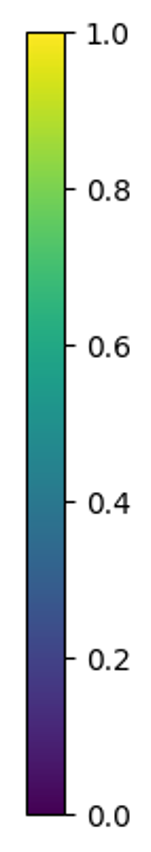}\\
    $\tau=1$ & $\tau=0.5$ & $\tau=0$ & GT-nearest & \\
    \end{tabular}
    \vspace{-0.3cm}
    \caption{Visualization of varying values of transmission coefficient $\tau$ in our global acoustic field. $\tau=1$ means all the sound energy will be transmitted to the other side of the wall, while $\tau=0$ represents no sound energy will be transmitted.}
    \vspace{-0.3cm}
\label{fig:vis_prior}
\end{figure*}

\noindent\textbf{Metrics.} On the RWAVS dataset, we follow AV-NeRF~\cite{AVNeRF} to utilise the magnitude distance (MAG)~\cite{xu2021visually} and envelope distance (ENV)~\cite{morgado2018self} as evaluation metrics. 
MAG measures the audio quality of the generated sound in the time-frequency domain after applying the short-time Fourier transform, while ENV measures it in the time domain. On the SoundSpaces dataset, we follow~\cite{AVNeRF, INRAS} to evaluate our method with the reverberation time (T60), acoustic parameter clarity (C50), and early decay time (EDT) metrics. T60 is the percentage error between the time it takes for the synthesised RIR to decay by 60 dB in the time domain with the ground truth T60 reverberation time. C50 describes the clarity and loudness of audio by quantifying the energy ratio between early reflections and late reverberation. EDT reflects people's perception of reverberation by focusing on the early reflections of impulse responses. For all of these metrics, lower is better. The detailed definitions of these metrics are provided in the supplementary materials.

\subsection{Results \& Ablation study }
We present the quantitative experimental results on the RWAVS dataset in Table~\ref{tab:rwavs}. Our model consistently outperforms all baselines across all environments. Specifically, we achieve an overall $7.2\%$ and $23.7\%$ reduction in the MAG metric compared to previous state-of-the-art audio-visual methods AV-NeRF~\cite{AVNeRF} and ViGAS~\cite{ViGAS}, respectively. This demonstrates that our approach can extract more comprehensive environmental information from visual inputs and efficiently integrate it into the neural acoustic field learning.~Table~\ref{tab:soundspaces} provides the quantitative results on the Soundsapces dataset. Compared to all previous methods, our approach achieves the best performance in all metrics. In particular, we obtain an overall $7.3\%$ reduction in T60 error compared to AV-NeRF~\cite{AVNeRF} and a $27.1\%$ reduction compared to INRAS~\cite{INRAS}, respectively. 
The consistent improvement further shows that our geometric priors can be useful for modeling impulse response. An example visual comparison of rendered audio is shown in Figure~\ref{fig:vis}.

\noindent\textbf{Ablation of Proposed Components.} 
We conducted an ablation study based on AV-NeRF on the RWAVS dataset. In Table~\ref{tab:rwavs_ablation}, ``w/o \textit{geo}, \textit{dir}'' uses AV-NeRF's default input (visual features, sound source, receiver location, and their relative direction). ``w/o \textit{dir}'' only adds our global-local acoustic field (\textit{geo}), showing improvements in all scenarios, demonstrating the effectiveness of our spatial geometric prior. 
``Ours - w/o \textit{dir}'' uses AV-NeRF's relative direction information and \textit{geo}, while ``Ours - full'' incorporates all proposed modules, achieving the best performance.

\noindent\textbf{Robustness to Reconstruction Quality.} 
Our method reconstructs wall layouts to ensure that transmission attenuation occurs at the correct locations in our global priors. To evaluate the robustness of our approach to 3D reconstruction quality, we visualize audio synthesis results in relation to scene reconstruction accuracy in Figure~\ref{fig:ablation_geo}. Specifically, we employ a vanilla NeRF~\cite{nerf} (Figure~\ref{fig:ablation_geo}.B.\textit{c}) and MonoSDF~[\textcolor{green}{61}] (Figure~\ref{fig:ablation_geo}.B.\textit{b}) to obtain 3D scene reconstructions of varying quality. As shown in Figure~\ref{fig:ablation_geo}.A, our audio synthesis performance is only slightly affected when the reconstruction error remains below 6.5 and degrades when the reconstructed room layouts become noisy.

\noindent\textbf{Contribution of Transmission Coefficient $\tau$.} 
We evaluate our method across four scenes with wall \textit{occ}lusions and four without, varying $\tau$ values. The average results are reported in Table~\ref{table:tau}. For single-room scenes, our method achieves a steady improvement with varying $\tau$ values, demonstrating the effectiveness of combining our distance-aware priors with direction-aware attentions. However, for multi-room scenes, setting $\tau = 1$ produces pure distance-aware priors for receivers in other rooms, resulting in complete ignorance of wall occlusions, while $\tau = 0$ uniformly assigns zero values to all points behind the wall, leading to undifferentiated prior knowledge between these points. Based on experiments, we achieve a balanced integration of distance and occlusion effects by setting $\tau =0.25$ for representative priors across receivers. A visualization of the impact of $\tau$ in our global acoustic field is shown in Figure~\ref{fig:vis_prior}.

\noindent\textbf{Spatial Effect.}
We report the Left-Right Energy (LRE) error~\cite{ViGAS} on the RWAVS dataset in Table~\ref{tab:eval_LRE}. The consistent decrease in all metrics, including the LRE error, shows that our direction-aware attention improves the spatial effect by providing informative binaural features.

\begin{table}[t]
\centering
\resizebox{0.95\linewidth}{!}{
\footnotesize
\begin{tabular}{l|cc|cc}
\toprule
\multirow{3}{*}{\quad Methods} & \multicolumn{2}{c|}{Single-room \scriptsize{(w/o \textit{occ})}} & \multicolumn{2}{c}{Multi-room \scriptsize{(w/ \textit{occ})}} \\ 
\cmidrule(lr){2-3} \cmidrule(lr){4-5}
& MAG$\downarrow$ & ENV$\downarrow$ & MAG$\downarrow$ & ENV$\downarrow$ \\
\midrule
AV-NeRF~[\textcolor{green}{25}] & 0.780 & 0.119 & 1.963 & 0.159 \\ 
Ours ($\tau=1$) & 0.672 & 0.113 & 1.902 & 0.158 \\
Ours ($\tau=0.5$) & 0.672 & 0.113 & 1.864 & 0.157 \\
\textbf{Ours} ($\bm{\tau=0.25}$) & \textbf{0.672} & \textbf{0.113} & \textbf{1.856} & \textbf{0.157} \\
Ours ($\tau=0$) & 0.673 & 0.113 & 1.871 & 0.157 \\
\bottomrule
\end{tabular}}
\vspace{-0.2cm}
\caption{Ablation study of transmission coefficient $\tau$.}
\vspace{-0.2cm}
\label{table:tau}
\end{table}
\begin{table}[t]
\centering

\resizebox{0.75\linewidth}{!}{
\footnotesize
\begin{tabular}{l|ccc}
\toprule
\multirow{3}{*}{\quad Methods} & \multicolumn{3}{c}{Overall} \\ 
\cmidrule(lr){2-4}
 & MAG$\downarrow$ & ENV$\downarrow$ & LRE$\downarrow$ \\
\midrule
Ours - w/o \textit{geo, dir} & 1.506 & 0.145 & 0.988 \\
Ours - w/o \textit{dir} & 1.450 & 0.143 & 0.982 \\
Ours - \textit{full} & \textbf{1.396} & \textbf{0.142} & \textbf{0.956} \\
\bottomrule
\end{tabular}}
\vspace{-0.2cm}
\caption{Performance comparison on the RWAVS dataset.}
\vspace{-0.2cm}
\label{tab:eval_LRE}
\end{table}

\section{Conclusion}
In this work, we introduced effective priors for the sound field learning, derived from distance-aware parametric sound propagation modeling and scene occlusions extracted from the input video. Our proposed direction-aware attention mechanism captures useful local features for binaural channels. Tested on the real dataset RWAVS and the synthetic dataset SoundSpaces, our approach outperforms existing works in audio generation.

\noindent\textbf{Limitations.} Following previous methods~\cite{AVNeRF}, our model is scene-specific and deals with a static sound source. Currently, the proposed priors focus solely on geometric aspects—distance, occlusion, and direction—which have proven effective for audio synthesis. We believe that expanding them to incorporate factors such as reverberation or time of flight is a promising direction for future research.

\clearpage
{
    \small
    \bibliographystyle{ieeenat_fullname}
    \bibliography{egbib}
}

\clearpage
\setcounter{page}{1}
\maketitlesupplementary

\section{Room Impulse Response Prediction}

On the SoundSpaces dataset, following~\cite{AVNeRF, INRAS}, we modify our framework to predict room impulse responses in the time domain instead of estimating acoustic masks. Specifically, we discard the mixture mask ${\bf m}_m$ and replace the output layers of MLPs to synthesize RIR signals $\mathbf{rir}_l$, $\mathbf{rir}_r$, rather than ${\bf m}^l_d$, ${\bf m}^r_d$, while leaving other components unchanged. The network is supervised by the $L_2$ distance between the predicted and ground-truth magnitudes, which are obtained by applying the Short-Time Fourier Transform (STFT) to the predicted and ground-truth impulse responses, with a setting of $512$ ffts, $128$ hop length, $512$ window length, and a hamming window specification.

Given the impulse responses for the left and right channels, $\mathbf{rir}_l$ and $\mathbf{rir}_r$, we can generate the binaural audio $a_t^*$ based on the source audio $a^*_s$ with temporal convolution:
\begin{equation}
    a_t^* = [a^*_s \otimes {\bf rir}_l, \enspace
    a^*_s \otimes {\bf rir}_r],
\end{equation}
where $\otimes$ denotes convolution operation.

\section{Evaluation Metrics}

\textbf{RWAVS Dataset.} We follow AV-NeRF~\cite{AVNeRF} to evaluate our method using metrics, such as the magnitude distance (MAG)~\cite{xu2021visually} and envelope distance (ENV)~\cite{morgado2018self} on the RWAVS dataset. 

MAG measures audio quality in the time-frequency domain and is defined as
\begin{equation}
    \mathrm{MAG}(\mathbf{m}_\mathrm{prd}, \mathbf{m}_\mathrm{gt}) = ||\mathbf{m}_\mathrm{prd} - \mathbf{m}_\mathrm{gt}||^2,
\end{equation}
where $\mathbf{m}_\mathrm{prd}$ and $\mathbf{m}_\mathrm{gt}$ are the predicted and ground truth magnitude, respectively. 

ENV describes audio quality in the time domain and is defined as
\begin{equation}
    \mathrm{ENV}(a_\mathrm{prd}, a_\mathrm{gt}) = ||\mathrm{hilbert}(a_\mathrm{prd}) - \mathrm{hilbert}(a_\mathrm{gt})||^2,
\end{equation}
where $a_\mathrm{prd}$ and $a_\mathrm{gt}$ are the predicted and ground truth audio, and $\mathrm{hilbert}$ denotes Hilbert transformation function~\cite{smith2008mathematics}.

\noindent\textbf{SoundSpaces Dataset.} Following~\cite{AVNeRF, INRAS}, we utilize the reverberation time (T60), acoustic parameter clarity (C50), and early decay time (EDT) as evaluation metrics on the SoundSpaces dataset. 

T60 is a crucial acoustic parameter that characterizes the reverberation effect of a room, defined as the time needed for a sound to decay by 60 decibels (dB). We calculate the percentage error of T60 by
\begin{equation}
    \mathrm{T60}(a_\mathrm{prd}, a_\mathrm{gt}) = \frac{|\mathrm{T60}(a_\mathrm{prd}) - \mathrm{T60}(a_\mathrm{gt})|}{\mathrm{T60}(a_\mathrm{gt})},
\end{equation}
where $a_\mathrm{prd}$ and $a_\mathrm{gt}$ are the predicted and ground truth impulse response. 

C50 is a measurement of audio clarity that quantifies the energy ratio between early reflections and late reverberation. The C50 distance is formatted as
\begin{equation}
    \mathrm{C50}(a_\mathrm{prd}, a_\mathrm{gt}) = |\mathrm{C50}(a_\mathrm{prd}) - \mathrm{C50}(a_\mathrm{gt})|.
\end{equation}

EDT reflects people's perception of reverberation by focusing on the early reflections of impulse responses. The EDT distance is formatted as
\begin{equation}
    \mathrm{EDT}(a_\mathrm{prd}, a_\mathrm{gt}) = |\mathrm{EDT}(a_\mathrm{prd}) - \mathrm{EDT}(a_\mathrm{gt})|.
\end{equation}

\section{Contribution of Occlusion Number $n$}

We split the receivers in a multi-room scene according to their number of occlusions $n$ to the sound source, and evaluate them separately. As shown in Table~\ref{tab:num_n}, our full model achieves improved performance by integrating the occlusion number $n$ to the sound source in the scene.

\begin{table}[!h]
\centering
\resizebox{1\linewidth}{!}{
\footnotesize
\begin{tabular}{l|cc|cc}
\toprule
& \multicolumn{2}{c|}{Region $n=1$} & \multicolumn{2}{c}{Region $n=2$} \\ 
\cmidrule(lr){2-3} \cmidrule(lr){4-5}
\quad Methods & MAG$\downarrow$ & ENV$\downarrow$ & MAG$\downarrow$ & ENV$\downarrow$ \\
\midrule
Ours - w/o \textit{n} & 1.912 & 0.144 & 0.390 & 0.076 \\
Ours - \textit{full} & \textbf{1.896} & \textbf{0.143} & \textbf{0.375} & \textbf{0.075} \\
\bottomrule
\end{tabular}}
\vspace{-0.1cm}
\caption{Ablation study of the occlusion number $n$.}
\label{tab:num_n}
\end{table}

\section{Additional Visualizations}

\begin{figure*}
    \centering
    \begin{subfigure}{0.64\linewidth}
    \includegraphics[width=\linewidth]{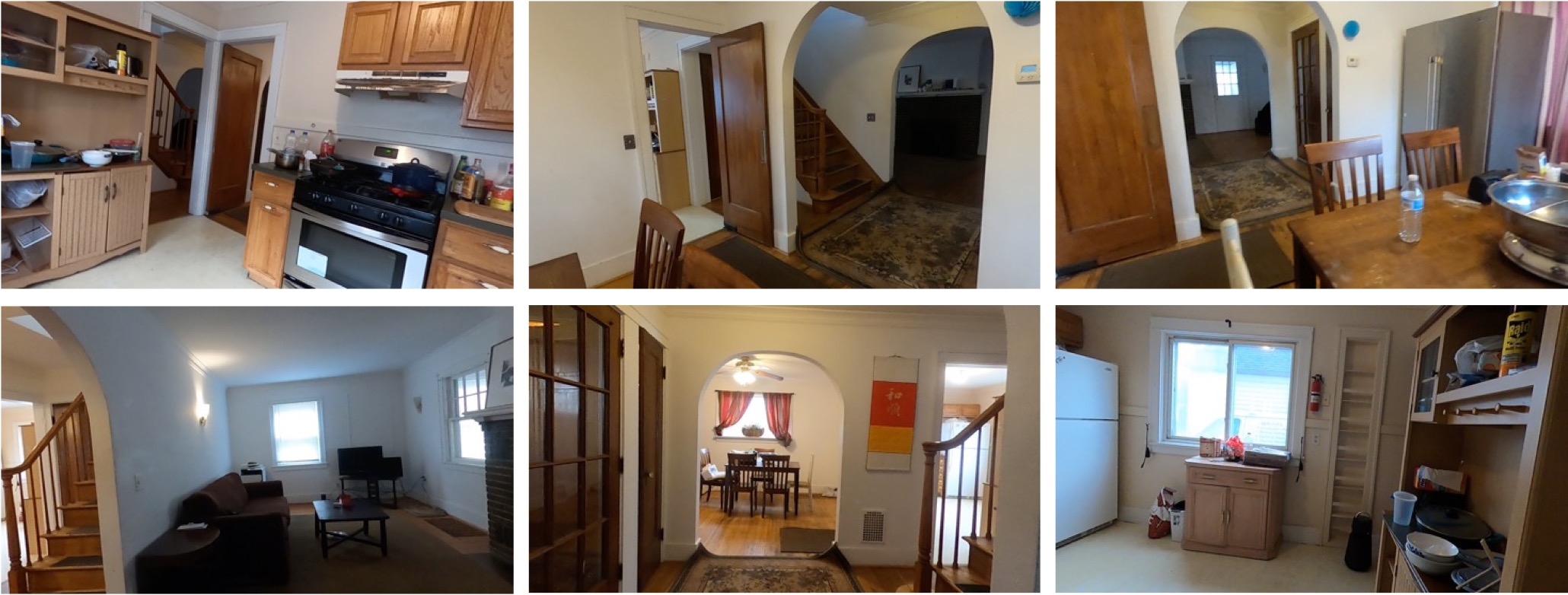}
    \caption{Example images of the input video}
    \end{subfigure}
    \hfill
    \begin{subfigure}{0.34\linewidth}
    \includegraphics[width=\linewidth]{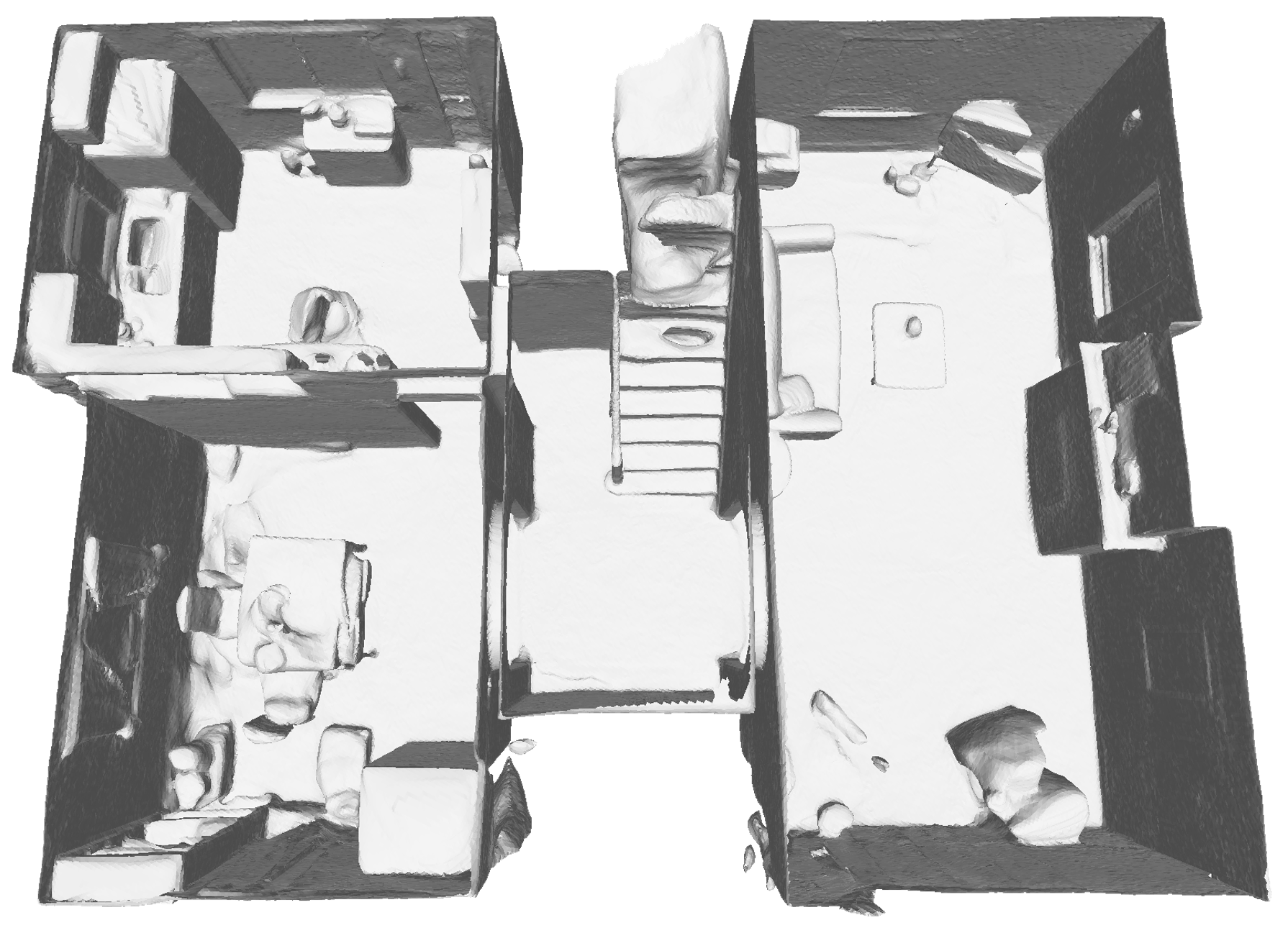}
    \caption{Reconstructed scene geometry}
    \end{subfigure}

     \vspace{0.5cm}
    \begin{subfigure}{1\linewidth}
    \centering
    \includegraphics[width=\linewidth]{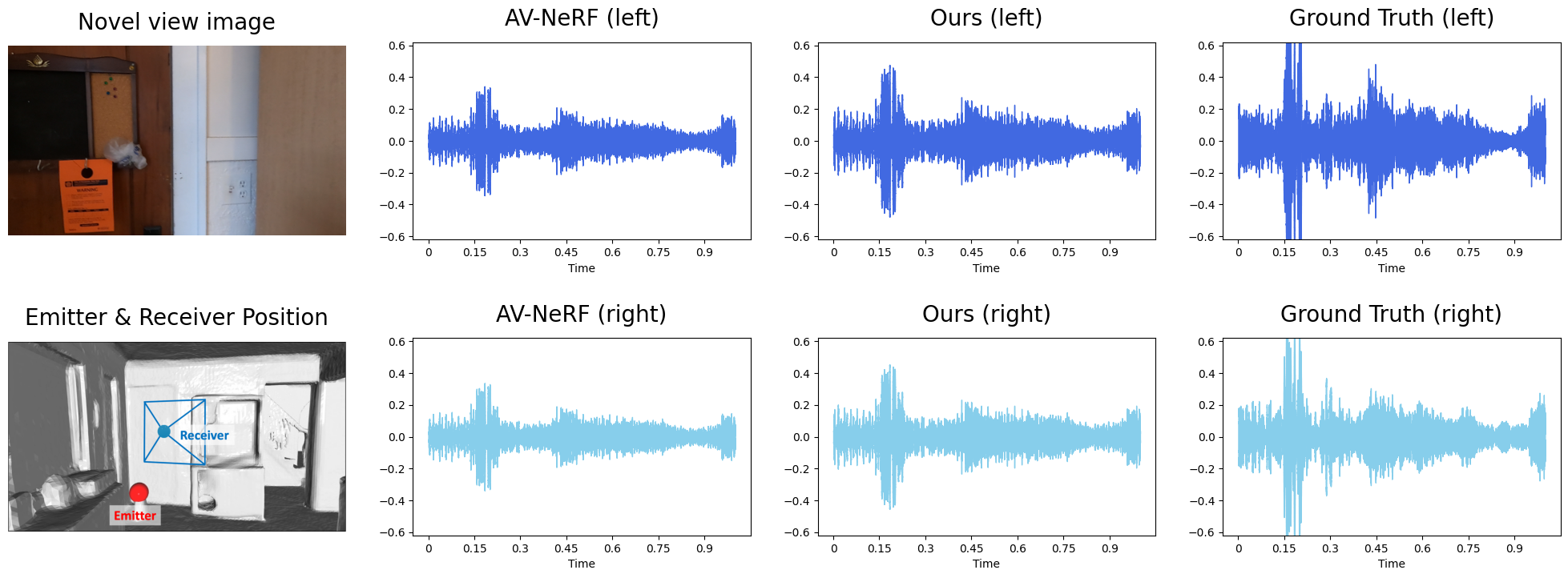}
    \caption{The receiver and emitter are located in the same room. Compared to AV-NeRF (MAG: 2.232; ENV:0.215), our method (MAG: 1.666; ENV:~0.188) synthesizes binaural audio with more accurate intensities.}
    \label{fig:quali_scene8_c}
    \end{subfigure}
    
    \vspace{0.5cm}
    \begin{subfigure}{1\linewidth}
    \centering
    \includegraphics[width=\linewidth]{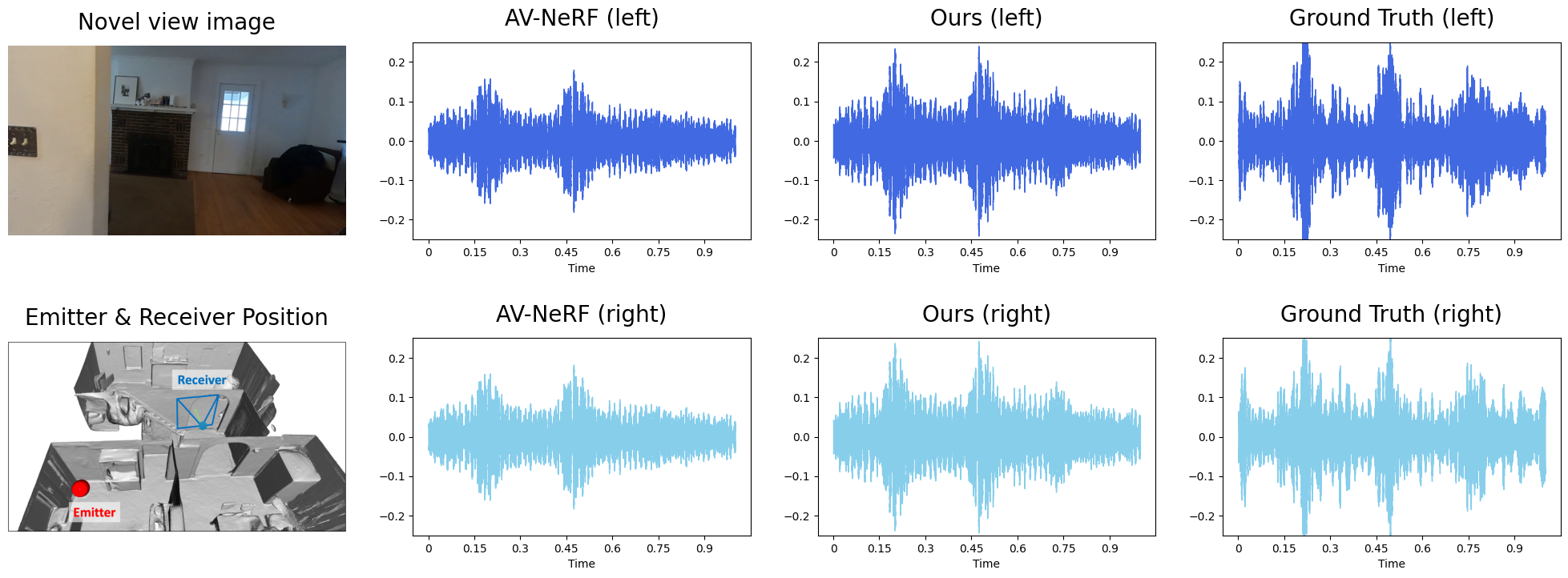}
    \caption{The receiver and emitter are located in different rooms. Compared to AV-NeRF (MAG: 0.573; ENV: 0.102), our method (MAG: 0.537; ENV:~0.098) generates audio with enhanced occlusion-aware attenuation.}
    \label{fig:quali_scene8_d}
    \end{subfigure}

    \vspace{0.3cm}
    \caption{Novel-view audio synthesis within a \textit{house} from the RWAVS dataset. (a) Example video images. (b) Scene geometry extracted from the input video. (c, d) Audio synthesis from two different receiver poses. Our approach generates superior binaural audio with (c) distance-aware and (d) occlusion-aware  effects than previous state-of-the-art method AV-NeRF~\cite{AVNeRF}.} 
    \label{fig:quali_scene8}
\end{figure*}

\begin{figure*}
    \centering
    \begin{subfigure}{0.64\linewidth}
    \includegraphics[width=\linewidth]{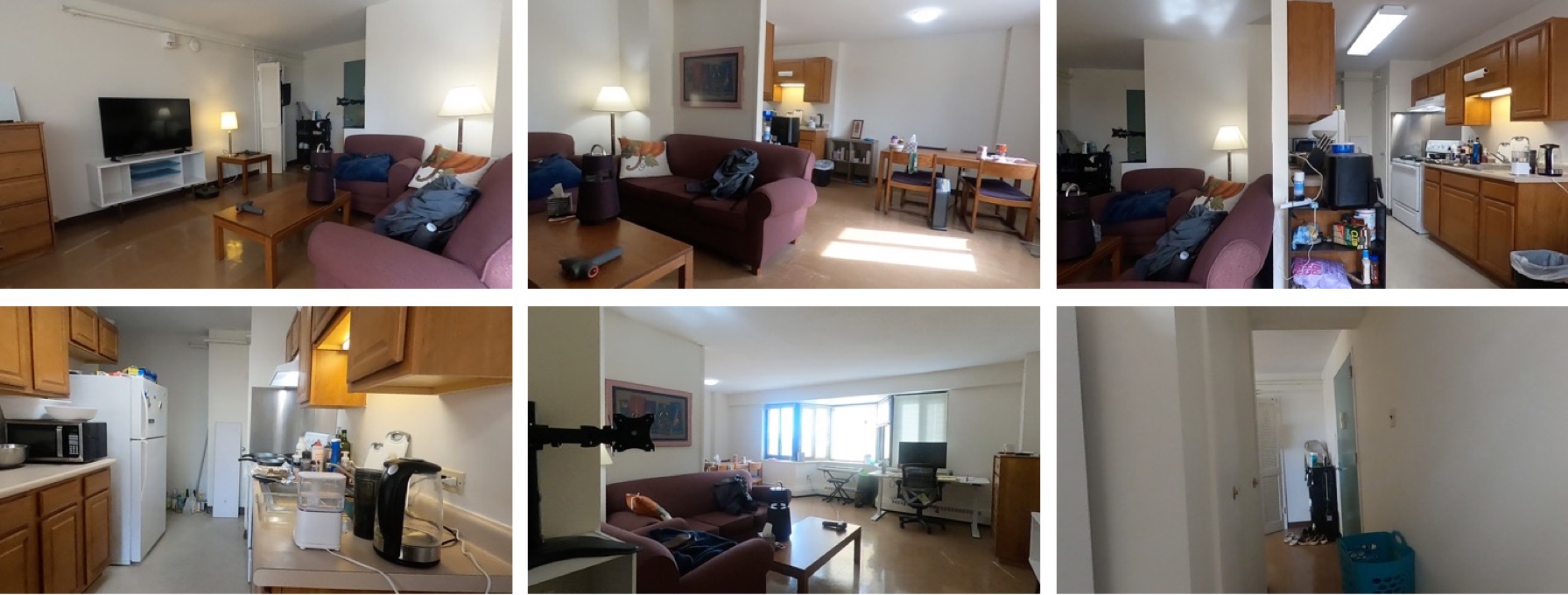}
    \caption{Example images of the input video}
    \end{subfigure}
    \hspace{0.2cm}
    \begin{subfigure}{0.32\linewidth}
    \includegraphics[width=\linewidth]{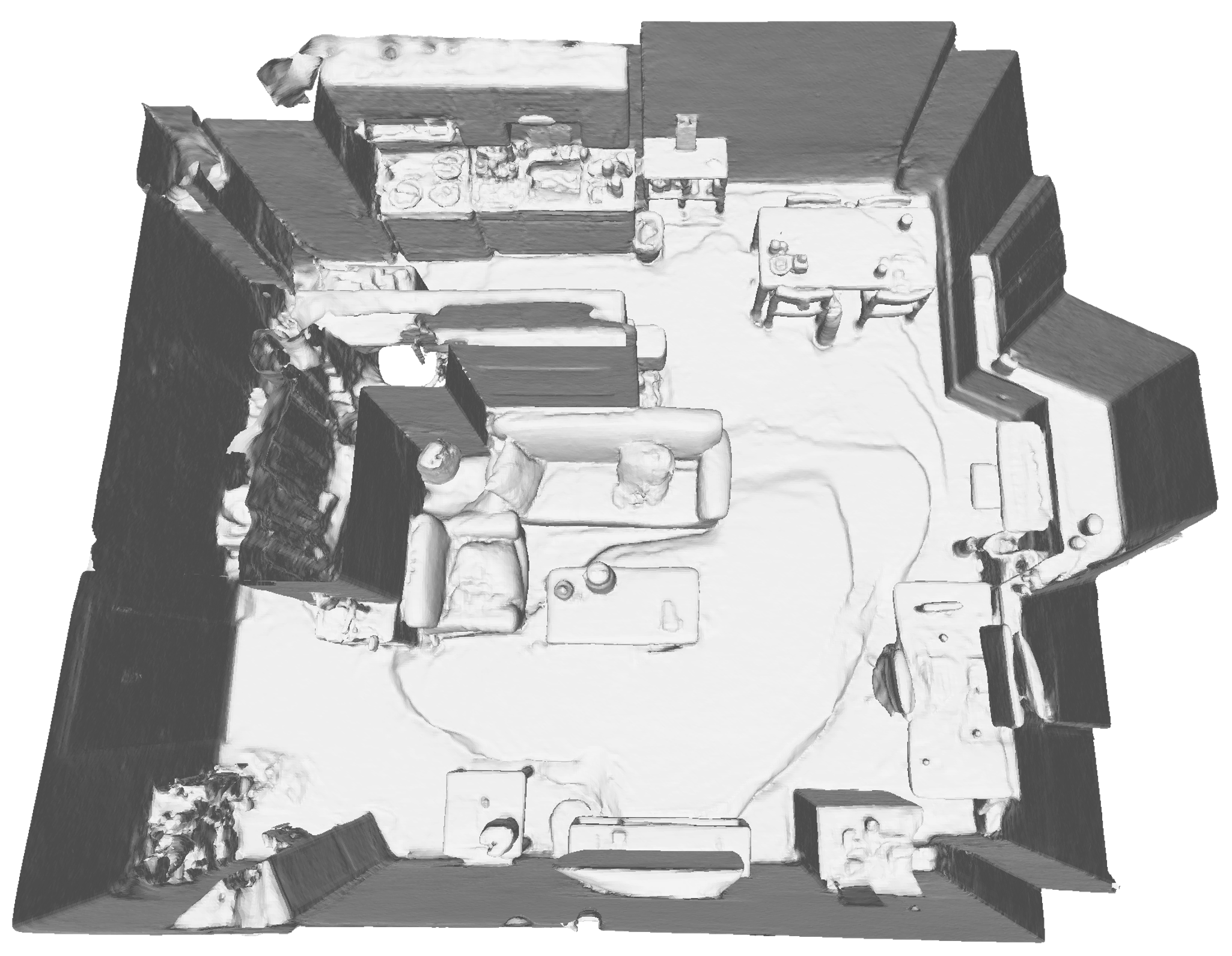}
    \caption{Reconstructed scene geometry}
    \end{subfigure}

    \vspace{0.5cm}
    \begin{subfigure}{1\linewidth}
    \centering
    \includegraphics[width=\linewidth]{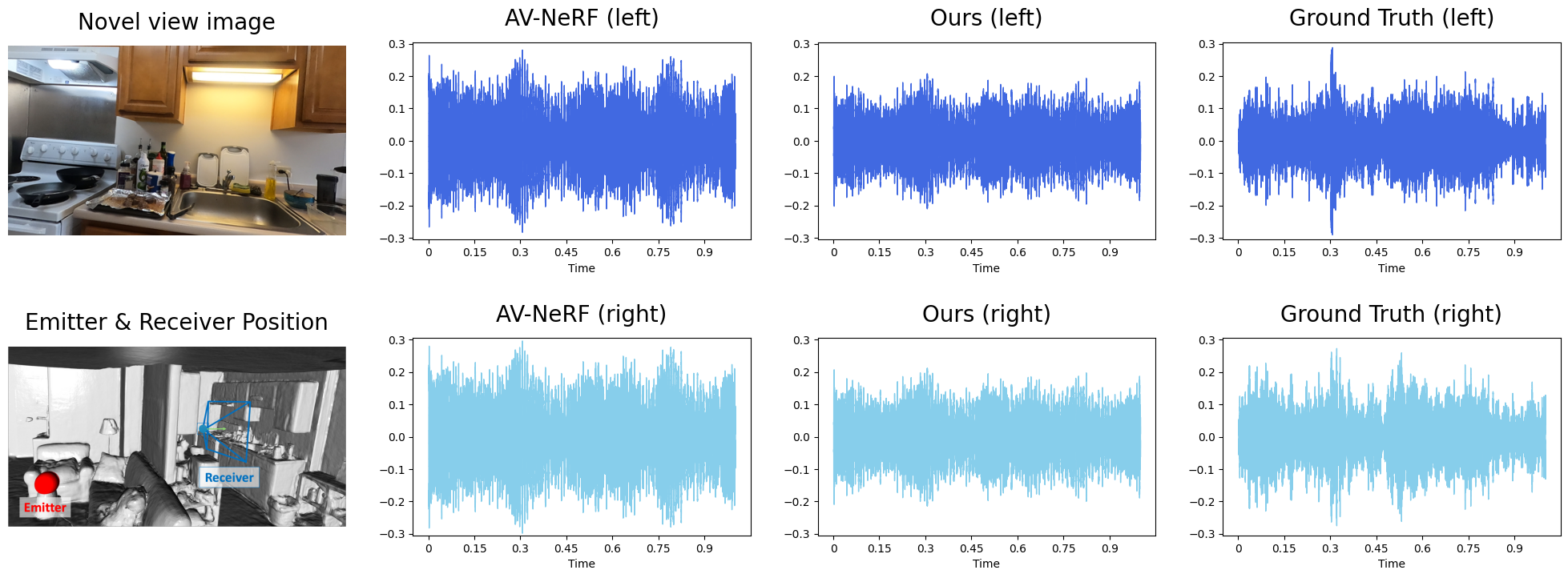}
    \caption{For the receiver, the emitter is blocked by a wall. Compared to AV-NeRF (MAG: 1.857; ENV: 0.159), our method (MAG: 0.726; ENV: 0.108) models the influence of walls on sound propagation and produces audio with more realistic attenuation.}
    \label{fig:quali_scene9_c}
    \end{subfigure}

    \vspace{0.5cm}
    \begin{subfigure}{1\linewidth}
    \centering
    \includegraphics[width=\linewidth]{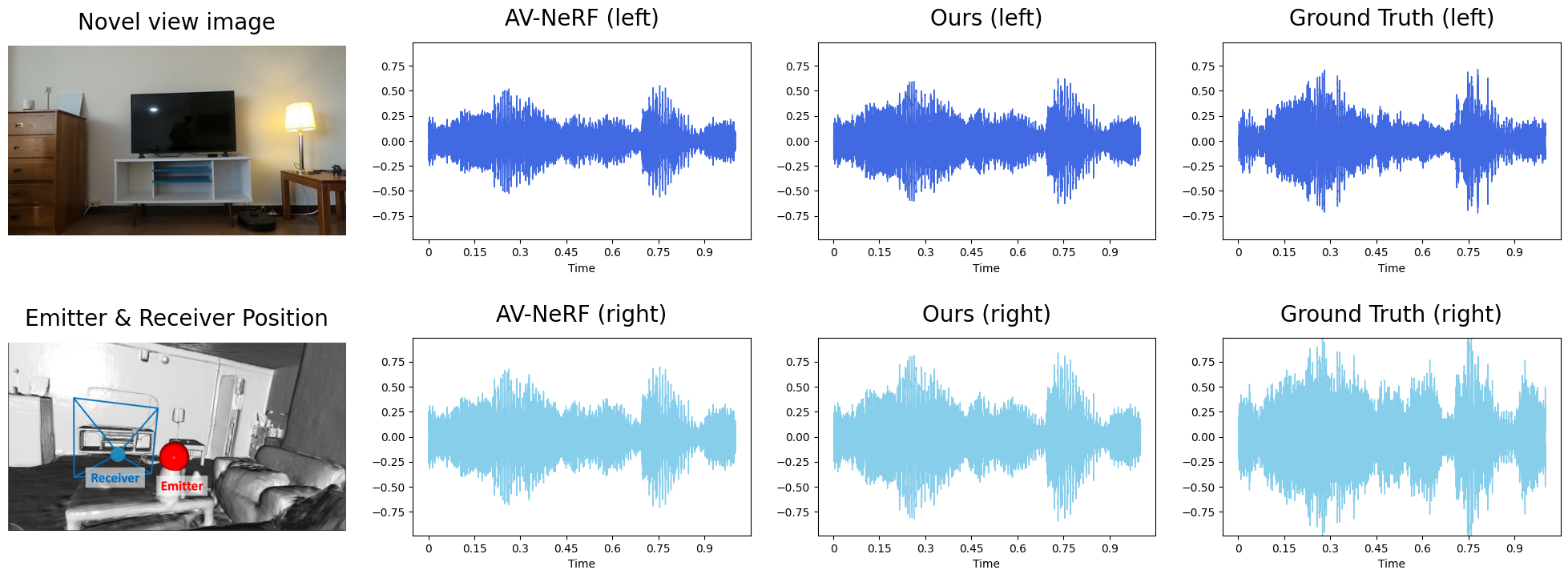}
    \caption{The receiver is to the left of the emitter. Compared to AV-NeRF (MAG: 4.636; ENV: 0.339), our method (MAG: 4.412; ENV: 0.334) improves spatial effects by synthesizing distinct audio in the left and right channels.}
    \label{fig:quali_scene9_d}
    \end{subfigure}

    \vspace{0.3cm}
    \caption{Novel-view audio synthesis within an \textit{apartment} from the RWAVS dataset. (a) Example video images. (b) Scene geometry extracted from the input video. (c, d) Audio synthesis from two different receiver poses. Our approach achieves better audio generation performance with (c) occlusion-aware and (d) direction-aware effects than previous state-of-the-art method AV-NeRF~\cite{AVNeRF}.}
    \label{fig:quali_scene9}
\end{figure*}

We provide additional qualitative comparisons of the synthesized audio within different scenes in Figure~\ref{fig:quali_scene8} and Figure~\ref{fig:quali_scene9}. For each scene, we present several example images of the input video, alongside the scene geometry that we reconstructed from the video frames. As shown in these figures, compared with AV-NeRF~\cite{AVNeRF}, our method demonstrates a superior ability to generate binaural audio with enhanced scene distance-aware (Figure~\ref{fig:quali_scene8_c}), occlusion-aware (Figure~\ref{fig:quali_scene8_d},~\ref{fig:quali_scene9_c}) and direction-aware (Figure~\ref{fig:quali_scene9_d}) effects from novel views in various scenarios.


\end{document}